\documentclass[10pt,twocolumn,letterpaper]{article}

\usepackage{cvpr}              
\usepackage{multirow}
\usepackage{makecell} 
\usepackage{amsthm}

\usepackage{graphicx}   
\usepackage{booktabs}   
\definecolor{mygreen}{RGB}{146, 199, 113} 
\definecolor{mypurple}{RGB}{146, 199, 113}
\usepackage[table]{xcolor}

\definecolor{cvprblue}{rgb}{0.21,0.49,0.74}
\usepackage[pagebackref,breaklinks,colorlinks,allcolors=cvprblue]{hyperref}

\def\paperID{ 2427} 
\def\confName{CVPR}
\def\confYear{2026}

\title{A Unified Framework for Knowledge Transfer in Bidirectional Model Scaling}

\author{Jianlu Shen\textsuperscript{\rm 1,2}\quad Fu Feng\textsuperscript{\rm 1,2}\quad Jiaze Xu\textsuperscript{\rm 1,2}\quad Yucheng Xie\textsuperscript{\rm 1,2}\quad Jiaqi Lv\textsuperscript{\rm 1,2}\thanks{Corresponding authors}\quad Xin Geng\textsuperscript{\rm 1,2}\footnotemark[1]\\
\textsuperscript{\rm 1}School of Computer Science and Engineering, Southeast University, Nanjing, China\\
\textsuperscript{\rm 2}Key Laboratory of New Generation Artificial Intelligence Technology and Its Interdisciplinary \\Applications (Southeast University), Ministry of Education, China\\
{\tt\small \{jlshen, fufeng, xujz, xieyc, jiaqi.lv, xgeng\}@seu.edu.cn}\\
}

\begin{document}
\maketitle
\begin{abstract}
Transferring pre-trained knowledge from a source model to a target model of a different architectural size is a key challenge for flexible and efficient model scaling. 
However, current parameter-space methods treat Small-to-Large (S2L) and Large-to-Small (L2S) scaling as separate, incompatible problems, focusing on parameter synthesis and selection, respectively.
This fragmented perspective has resulted in specialized tools, hindering a unified, bidirectional framework.
In this paper, we propose BoT (Bidirectional knowledge Transfer), the first size-agnostic framework to unify S2L and L2S scaling.
Our core insight is to treat model weights as continuous signals, where models of different sizes represent distinct discretizations of the transferable knowledge.
This multi-resolution perspective directly casts S2L and L2S scaling as the signal processing operations of upsampling and downsampling, naturally leading to the adoption of the Discrete Wavelet Transform (DWT) and its Inverse (IDWT).
BoT leverages the recursive nature of wavelets, using the decomposition level as a dynamic scaling factor to bridge disparate model sizes in a parameter-free and computationally efficient manner. 
Extensive experiments on DeiT, BERT, and GPT demonstrate significant pre-training FLOPs savings (up to 67.1\% for S2L, 52.8\% for L2S) and state-of-the-art performance on benchmarks like GLUE and SQuAD.

\end{abstract}

\vspace{-0.1in}
\section{Introduction}
\label{sec:intro}

Modern deep learning relies heavily on ``model zoos" that offer pre-trained knowledge in discrete, fixed-size architectures (\eg, -Base, -Large).
This pre-train and fine-tune paradigm, while effective, suffers from a core limitation: knowledge transfer is often strictly restricted to identical source and target models~\cite{romero2014fitnets,yang2025foundation}.
This tight coupling between the learned knowledge and the fixed architectural dimensions makes transferring knowledge between mismatched models non-trivial, creating a significant and costly barrier to flexible model scaling.

\begin{figure}[tb]
    \centering
    \includegraphics[width=\columnwidth]{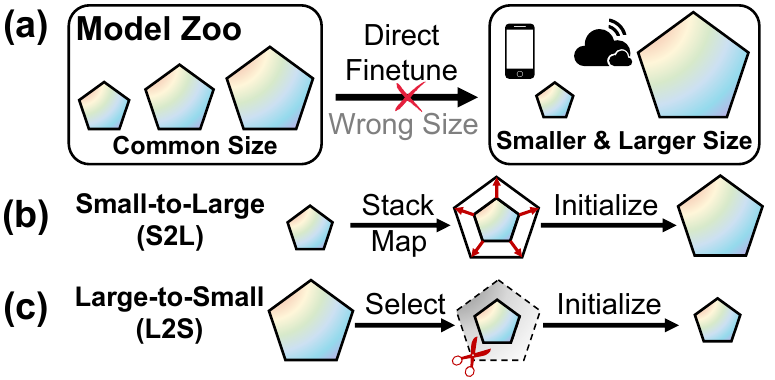}
    \vspace{-0.28in}
    \caption{Conceptual illustration of knowledge transfer challenges between mismatched model sizes. (a) Direct fine-tuning from model zoos is often unviable due to size mismatches. (b) Small-to-Large (S2L) methods initialize by mapping or stacking weights from a small model. (c) Large-to-Small (L2S) methods initialize by selecting weights from a large model. }
    \label{fig:motivation}
\vspace{-0.27in}
\end{figure}

This barrier to scaling is twofold.
First, driven by empirically validated Scaling Laws~\cite{kaplan2020scaling}, the primary strategy for achieving state-of-the-art performance is training ever-larger models~\cite{chen2022scaling,floridi2020gpt}.
This process is computationally prohibitive when starting from scratch~\cite{schwartz2020green}, as it fails to leverage the knowledge embedded in existing, smaller models. 
This inefficiency drives the need for efficient Small-to-Large~(S2L) model growth~\cite{wang2023learning}, which seeks to reuse the parameters of a smaller model to construct a strong initial state for a larger target model, thereby accelerating its convergence.
Second, while the massive and general-purpose large models are successfully trained, their prohibitive inference costs and large memory footprints make them impractical for deployment in many real-world applications~\cite{sanh2019distilbert}.
This creates a critical need for Large-to-Small~(L2S) model adaption to efficiently transfer the generalizable knowledge into resource-tailored architectures.

To address these challenges, recent research focuses on reusing knowledge \emph{from the parameter space} of a source model to optimally initialize a target architecture. However, existing methods have bifurcated into methodologically incompatible trajectories. As shown in Figure~\ref{fig:motivation}, S2L transfer is framed as a \emph{parameter synthesis} problem, employing techniques like layer duplication~\cite{chen2021bert2BERT} or trainable mapping functions~\cite{wang2023learning, pan2023reusing} that incur extra training overhead. Conversely, L2S transfer is treated as a \emph{parameter selection} problem, relying on training-free heuristics~\cite{xu2023initializing} to extract weight subsets. This dichotomy has locked the field into developing specialized, ad-hoc tools, obscuring the fact that both directions are fundamentally identical problems of bidirectional model scaling.
The \emph{learngene} paradigm~\citep{feng2023genes, feng2024wave, feng2024transferring, xie2024kind, xie2025divcontrol, xie2024fine} provides a compelling theoretical lens to resolve this fragmentation. It posits that pre-trained models encapsulate a condensed, intrinsic core of knowledge---akin to biological genes---that is inherently decoupled from specific architectural dimensions. If this ``genetic" representation can be effectively isolated and inherited, models of any scale could acquire a robust foundation for rapid downstream adaptation. This raises a critical question: how can we materialize a unified, \emph{size-agnostic} learngene to transcend current disjointed operations, seamlessly unifying S2L and L2S scaling within the parameter space?


In this paper, we propose \textbf{B}idirectional kn\textbf{o}wledge \textbf{T}ransfer~(BoT), the first algorithmic answer to this question. 
Our core insight is to \emph{treat model weights as continuous signals}, since recent advances
provide growing evidence that the parameter space of well-performing models is not random but highly structured, populating a low-dimensional manifold~\cite{schrholt22hyper,soro25diffusion}. 
We thereby hypothesize that this underlying signal represents the generalizable knowledge, and that the fundamental low-frequency spectrum inherently encapsulates the \emph{learngene}. Models of different architectural sizes are simply different resolution discretizations of this same signal.
A small model, by nature of its limited capacity, is forced to capture the low-resolution, global approximation of the knowledge, akin to a blurry thumbnail of an image, and a larger model has the capacity to refine this by adding high-resolution, task-dependent details.
This multi-resolution perspective conceptualizes L2S and S2L scaling as downsampling and upsampling operations, leading us naturally to the Discrete Wavelet Transform (DWT)~\cite{heil1989continuous} and its Inverse~(IDWT).
This DWT/IDWT method is intrinsically size-agnostic by leveraging the recursive, multi-level nature of the wavelet transform. 
The number of decomposition levels serves as a dynamic scaling factor to transfer this learngene across mismatched model sizes.
Specifically, for L2S adaptation, we apply DWT to condense the large model's knowledge into a compact learngene, comprising low-frequency coefficients scaled precisely to the target size. Conversely, for S2L expansion, we treat the small model's weights as the foundational learngene. By padding the absent high-frequency details with zeros and applying IDWT, we seamlessly reconstruct the larger architecture. This bidirectional mechanism guarantees a \emph{parameter-free} and \emph{computationally efficient} initialization across arbitrary model scales.

We demonstrate BoT's effectiveness through extensive experiments on diverse architectures, including Vision Transformers~(DeiT), encoder-based models~(BERT), and decoder-based models~(GPT).
For S2L, BoT's upsampling initialization allows large target models to converge faster, saving up to 67.1\% (BERT), 58.3\% (GPT), and 22.0\% (DeiT) in pre-training FLOPs compared to training from scratch. 
For L2S, the small models initialized by BoT's downsampling also converge significantly faster, yielding FLOPs savings of up to 52.8\% (BERT), 39.0\% (DeiT), and 31.0\% (GPT) over their from-scratch counterparts.
Furthermore, models inheriting the BoT learngene demonstrate the state-of-the-art performance on a wide range of downstream tasks, including visual tasks and language benchmarks like the GLUE~\cite{wang2018glue} and SQuAD~\cite{rajpurkar2016squad,rajpurkar2018know}. 
Visualizations confirm that BoT preserves the intrinsic, structured patterns observed in original pre-trained weights~\cite{trockman2023mimetic,xu2023initializing}

\section{Related Work}
\label{sec:related_work}

\paragraph{Model Initialization}
\label{subsec:model_initialization}
Effective model initialization has evolved from foundational \textit{from-scratch} statistical methods~\cite{glorot2010understanding, he2015delving} to the now-dominant pre-train and fine-tune paradigm. This paradigm, however, is primarily effective only when source and target architectures are identical, giving rise to significant challenges for cross-architecture initialization. 
The first, L2S transfer, focuses on adapting large pre-trained models for resource-constrained hardware. A prominent training-free approach, Weight Selection (WS)~\cite{xu2023initializing}, directly samples parameters from the larger model, but this heuristic-based selection risks disrupting learned structural patterns. The second, S2L transfer, seeks to accelerate the training of larger models by initializing from smaller ones. Initial approaches, like bert2BERT~\cite{chen2021bert2BERT}, utilized naive parameter stacking to construct the larger model. The recent approaches have shifted towards complex, trainable mappings, such as the linear operators of LiGO~\cite{wang2023learning} or the multi-linear operations of Mango~\cite{pan2023reusing}. While effective, these learnable methods introduce additional computational overhead during the transfer process. In contrast to these specialized approaches, relying on either heuristic selection or costly trainable mappers, our method, BoT, provides a unified, and principled non-trainable solution that efficiently handles both transfer directions.

\vspace{-0.17in}

\begin{figure*}[tb]
    \centering
    \includegraphics[width=0.98\textwidth]{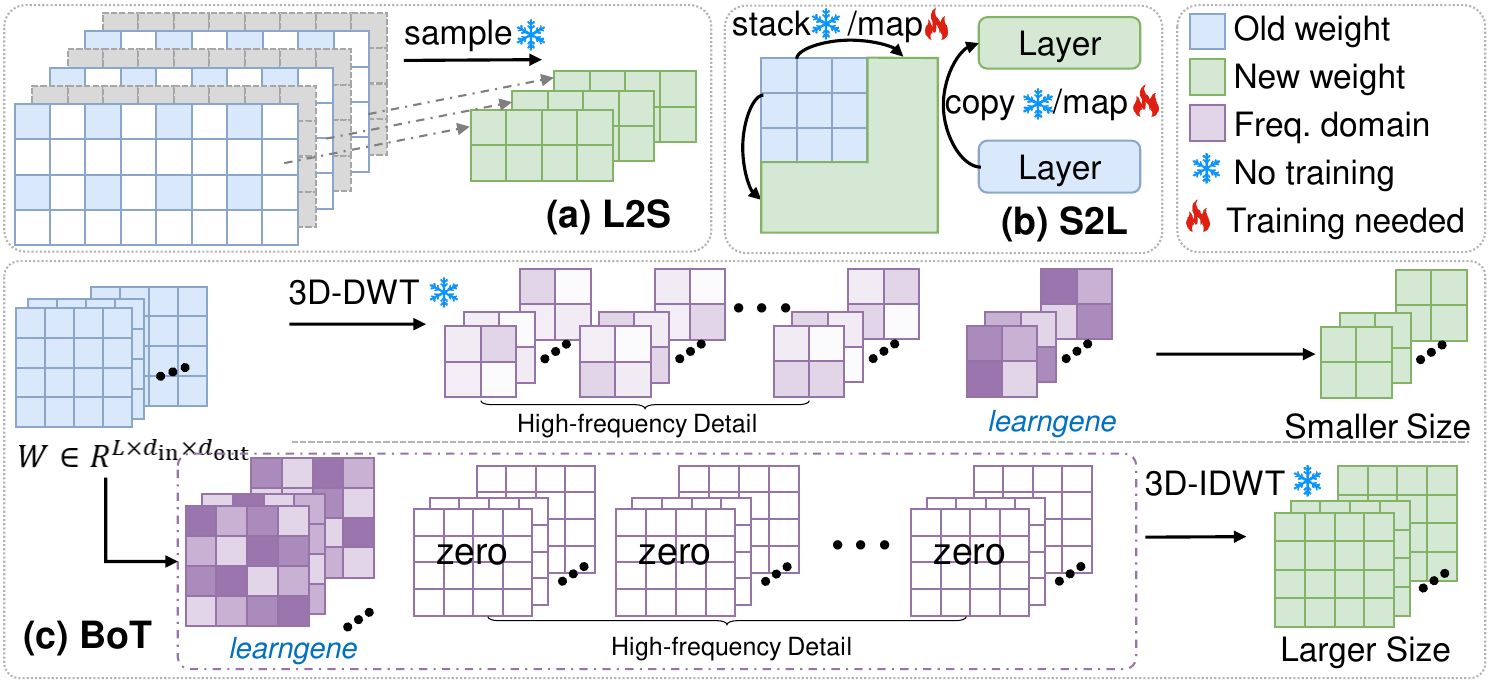}
    \vspace{-0.10in}
    \caption{Comparison of cross-architecture initialization frameworks. (a) Previous Large-to-Small (L2S) methods rely on heuristic sampling or layer selection. (b) Previous Small-to-Large (S2L) methods use direct copying or trainable mapping functions. (c) Our proposed BOT framework unifies both directions through a parameter-free, frequency-domain transformation using the 3D-DWT and IDWT. }
    \label{fig:framework}
    \vspace{-0.19in}
\end{figure*}

\paragraph{Frequency Analysis in Networks}
\label{subsec:related_freq}
The application of frequency-domain analysis in deep learning has primarily followed two approaches. The first, more established approach, applies frequency transforms to the input data or intermediate feature representations~\cite{gueguen2018faster, xu2020learning, rao2021global, liu2018frequency, scribano2023dct}. Notable examples include JPEG-based Convolutional Neural Networks~\cite{wallace1991jpeg, raid2014jpeg} and F-Net~\cite{lee2021fnet}, which replaces self-attention with a discrete fourier transform. The second approach applies transforms directly to network parameters. For instance, Ulicny \etal~\cite{ulicny2022harmonic, ulicny2021tensor} employed one or two dimension Discrete Cosine Transform (1D/2D-DCT) for model compression via weight pruning. However, these methods are all fundamentally \textit{intra-architecture}, designed to optimize efficiency or compress a single model. To our knowledge, we are the first to apply a 3D wavelet transform to model parameters, repurposing it from data analysis or pruning into a principled mechanism for \textit{cross-architecture} initialization. This novel application enables bidirectional, arbitrary-sized knowledge transfer, a problem unaddressed by prior frequency methodologies.

\section{Methods}
\label{sec:methods}

We propose BoT, a novel \textbf{B}idirectional kn\textbf{o}wledge \textbf{T}ransfer framework that leverages the wavelet transform as an initialization method for models of varying sizes. Our core insight is to repurpose the wavelet transform's fundamental operations, decomposition and reconstruction, as a general method for model scaling. 
Specifically, wavelet decomposition (DWT) facilitates Large-to-Small (L2S) transfer, while its inverse, wavelet reconstruction (IDWT), facilitates Small-to-Large (S2L) transfer.

\subsection{Preliminaries}
\label{subsec:preliminaries}

The Discrete Wavelet Transform (DWT) is a reversible mathematical tool that decomposes a signal into frequency components. The characteristics of this process are determined by a chosen wavelet family~\cite{lee2019pywavelets} (\eg, Haar), which defines a set of low-pass and high-pass filters. While traditionally applied to 1D signals or 2D images, we employ its 3D variant to process the structured weight parameters of neural networks. This inherent duality of decomposition and reconstruction is the foundation of BoT. Further mathematical details are provided in Appendix~\ref{app:dwt} and~\ref{app:sec_wavelet_family}.

\textbf{Decomposition (3D-DWT).} 
For a given 3D input ${W}$ of size $I \times J \times K$, the DWT separates it into two distinct parts: a single low-frequency approximation sub-band (${cA}$) and seven high-frequency detail sub-bands ($\{{cD}_m\}_{m=1}^7$). The transform operates by applying low-pass ($\Phi_d$) and high-pass ($\Psi_d$) filters along all three axes, each followed by down-sampling by a factor of $f$ (typically $f=2$). This down-sampling mechanism directly contracts the tensor's spatial resolution, yielding an approximation sub-band ${cA}$ of approximately $(\frac{I}{f} \times \frac{J}{f}\times \frac{K}{f})$. The ${cA}$ sub-band captures the global, low-resolution essence of the original ${W}$ in a compact form. The approximation coefficients are formally obtained by applying the low-pass operators: \begin{equation} 
\label{eq:dwt_3d_approx} 
{cA} = \Phi_k(\Phi_j(\Phi_i({W}))). \end{equation} 
The seven detail sub-bands are formed by other operator combinations and store the high-frequency details. For example, ${cD}_1 = \Psi_k(\Phi_j(\Phi_i({W}))).$

\textbf{Reconstruction (3D-IDWT).} In contrast, the IDWT is the synthesis operation that reverses DWT decomposition. It takes the coefficient sub-bands as input and, through up-sampling by $f$ and filtering, can perfectly reconstruct the original, full-sized input ${W}$: 
\begin{equation} 
\label{eq:idwt_3d} 
{W} = \text{IDWT}_{\text{3D}}({cA}, \{{cD}_m\}_{m=1}^7). 
\end{equation} 
Crucially, the IDWT can synthesize a full-sized output even if the high-frequency detail coefficients $\{{cD}_m\}$ are set to zero. This allows a larger structure ${W}$ to be reconstructed purely from the low-frequency ${cA}$ band.

\subsection{L2S Transfer via Wavelet Decomposition}
\label{subsec:Large-to-Small}

Our L2S transfer leverages the DWT's decomposition capability to extract a compact \emph{learngene} from a larger pre-trained source, efficiently initializing a smaller target model. The process is as follows:

\textbf{Parameter Consolidation.} First, we organize the pre-trained model's parameters into function-specific groups for establishing the structural basis for learngene extraction. This consolidation is particularly effective for modern architectures like the Transformer (\eg, DeiT~\cite{touvron2021training}, BERT~\cite{devlin2018bert}, GPT~\cite{floridi2020gpt}), which are built by stacking numerous, structurally-identical layers or blocks. We therefore use the Transformer as our canonical example. For a model with $L_{\text{src}}$ layers, we construct each group by stacking the corresponding 2D weight matrices from all layers, forming a 3D weight module. For instance, the query, key, and value projection weights from all $L_{\text{src}}$ layers are consolidated into a single module ${W}_{qkv}$. This consolidation is repeated for other principal components, such as the attention output projections (${W}_{o}$) and the two feed-forward network (FFN) layers (${W}_{f1}, {W}_{f2}$). The complete set of consolidated source parameters is thus:
\vspace{-0.08in}
\begin{equation} \label{eq:consolidation} \Omega_{\text{src}} = \{{W}_{qkv}, {W}_{o}, {W}_{f1}, {W}_{f2} \} \in \mathbb{R}^{L_{\text{src}} \times d_{\text{in, src}} \times d_{\text{out, src}}}, 
\vspace{-0.08in}
\end{equation} 
where $d_{\text{in, src}}$ and $d_{\text{out, src}}$ denote the input and output dimensions of the respective linear layers.



\textbf{Learngene Extraction and Initialization.} Second, we apply the 3D-DWT to each source module ${W}_{\text{src}} \in \Omega_{\text{src}}$, selecting a decomposition level (\ie, down-sampling factor $f$) that precisely matches the dimensions of the smaller target module ${W}_{\text{tgt}}$. The resulting low-frequency approximation sub-band, ${cA}_{\text{src}}$—formally derived via low-pass filter operators as in Equation~(\ref{eq:dwt_3d_approx})—acts as the condensed learngene. We directly initialize the target module by inheriting this core (${W}_{\text{tgt}} = {cA}_{\text{src}}$).

By inheriting this low-frequency learngene, the target model seamlessly retains the source model's core structural properties. This genetic transfer establishes a highly structured and informed starting point, significantly outperforming random initialization or heuristic sampling.

\subsection{S2L Transfer via Wavelet Synthesis}
\label{subsec:Small-to-Large}

Conversely, the S2L task leverages the IDWT's synthesis capability to expand a smaller pre-trained model, acting as the learngene, into a larger target architecture.

\textbf{Inheriting the Learngene. }We designate the pre-trained small model's consolidated weights, ${W}_{\text{src}}$, as the \textit{learngene} for the synthesis. This is implemented by setting the IDWT's approximation coefficients ${cA}$: ${cA} = {W}_{\text{src}}$.

\textbf{Synthesizing the Target Model.} Next, we set all seven high-frequency detail coefficients $\{{cD}_m\}$ to zero. The IDWT is then applied using the source model's weights as the low-frequency band (${cA} = {W}_{\text{src}}$) and these zeroed detail bands. This synthesizes a full-resolution weight module ${W}_{\text{tgt}}$ that matches the larger target dimensions:
\vspace{-0.08in}
\begin{equation} 
\label{eq:lowpass_recon_method}
{W}_{\text{tgt}} = \text{IDWT}{\text{3D}}({W}_{\text{src}}, {O_1}, \dots, {O_7}),
\vspace{-0.1in}
\end{equation}
where ${\{{O}_i\}_{i=1}^7}$ represents zero tensors of the required shapes. This operation effectively provides the target model with a stable, coherent foundation based entirely on the source's learned representations.

\section{Experiments}
\label{sec:experiments}

\subsection{Experimental Setup} 
\label{sec:setup}




\textbf{Comparison with Recent Advances.} BoT offers distinct advantages over existing cross-architecture knowledge transfer methods. While promising, they typically specialize in either model adaptation (Large-to-Small) or expansion (Small-to-Large) and differ significantly in their operational mechanisms. Table~\ref{tab:comparison} provides a qualitative comparison of BoT against the recent approaches.

\begin{table}[t]
\centering
\setlength{\tabcolsep}{0.9 mm}
\caption{Qualitative comparison of different cross-architecture initialization methods. ``No-Trainability" refers to whether the method itself directly transfers knowledge without introducing learnable parameters for the transfer process.}
\label{tab:comparison}
\vspace{-0.07in}
\resizebox{0.46\textwidth}{!}{
\begin{tabular}{@{}lccc@{}}
\toprule
Methods & Large-to-Small & {Small-to-Large} & {No-Trainability} \\ \midrule
WS~\cite{xu2023initializing} & \checkmark & $\times$ & \checkmark \\
bert2BERT~\cite{chen2021bert2BERT} & $\times$ & \checkmark & \checkmark \\
LiGO~\cite{wang2023learning} & $\times$ & \checkmark & $\times$ \\
Mango~\cite{pan2023reusing} & $\times$ & \checkmark & $\times$ \\
 \midrule
{BoT} & \checkmark & \checkmark & \checkmark \\ \bottomrule
\end{tabular}
}
\vspace{-0.19in}
\end{table}

The Large-to-Small (L2S) task is challenging, requiring the extraction of essential knowledge from large models to small target models. While optimization strategies like Knowledge Distillation (KD)~\cite{hinton2015distilling} can achieve this, it incurs a substantial, cumulative computational cost from repeated teacher inferences at every training step. In contrast, training-free approaches serve as an efficient, one-time initialization alternative, like Weight Selection (WS)~\cite{xu2023initializing}, shown in Figure~\ref{fig:framework}~(a), rely on parameter sub-sampling from the larger model. However, this heuristic risks disrupting the structural integrity of learned representations, as the selection process overlooks the crucial inter-dependencies established between parameters.

\begin{figure*}[tb]
    \centering
    \subfloat[DeiT-B $ \rightarrow$ DeiT-S\label{fig:deit_s}]{%
        \includegraphics[width=0.33\textwidth]{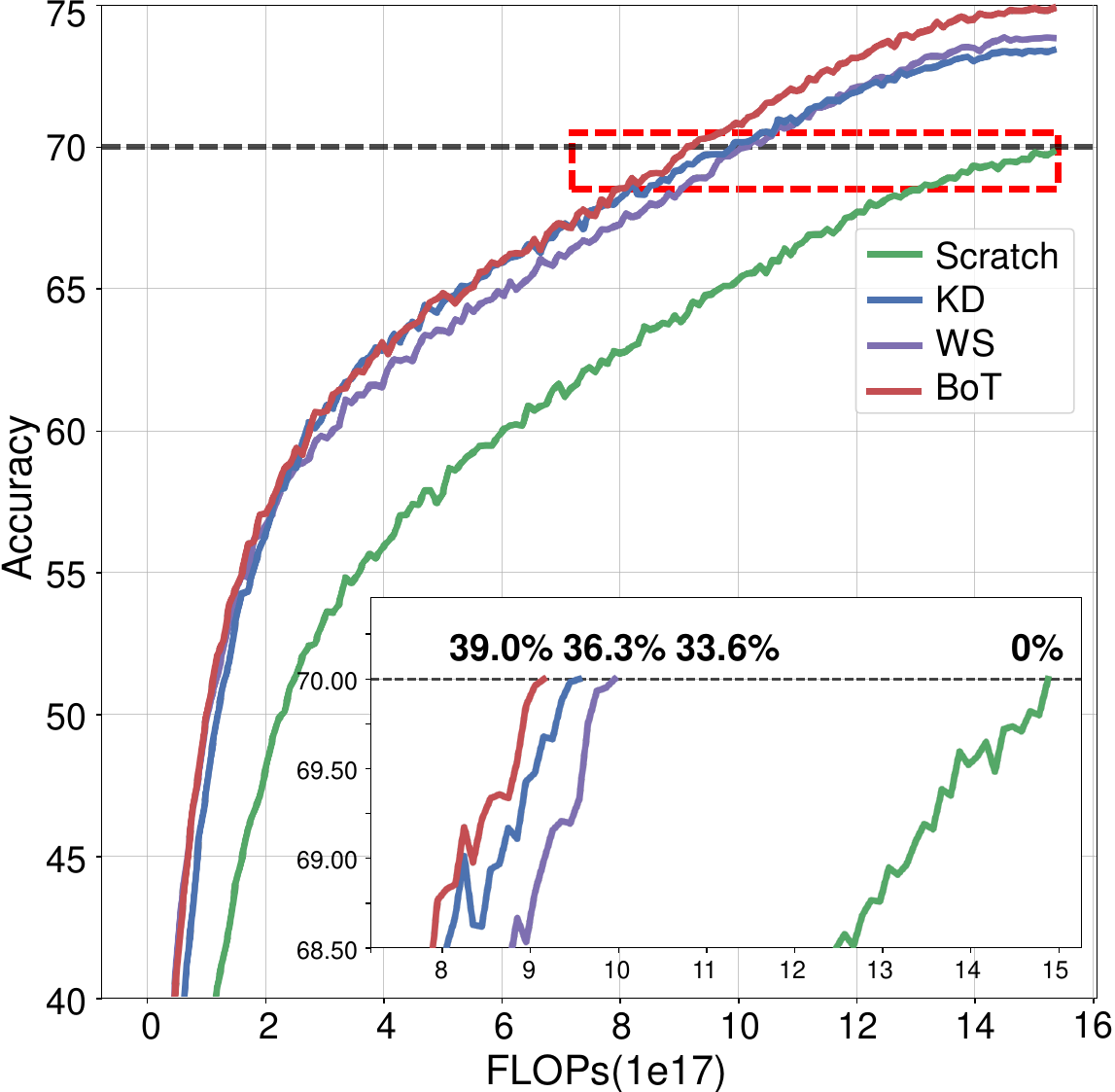}%
    }
    \hfill
    \subfloat[BERT-B $ \rightarrow$ BERT-S\label{fig:bert_s}]{%
        \includegraphics[width=0.33\textwidth]{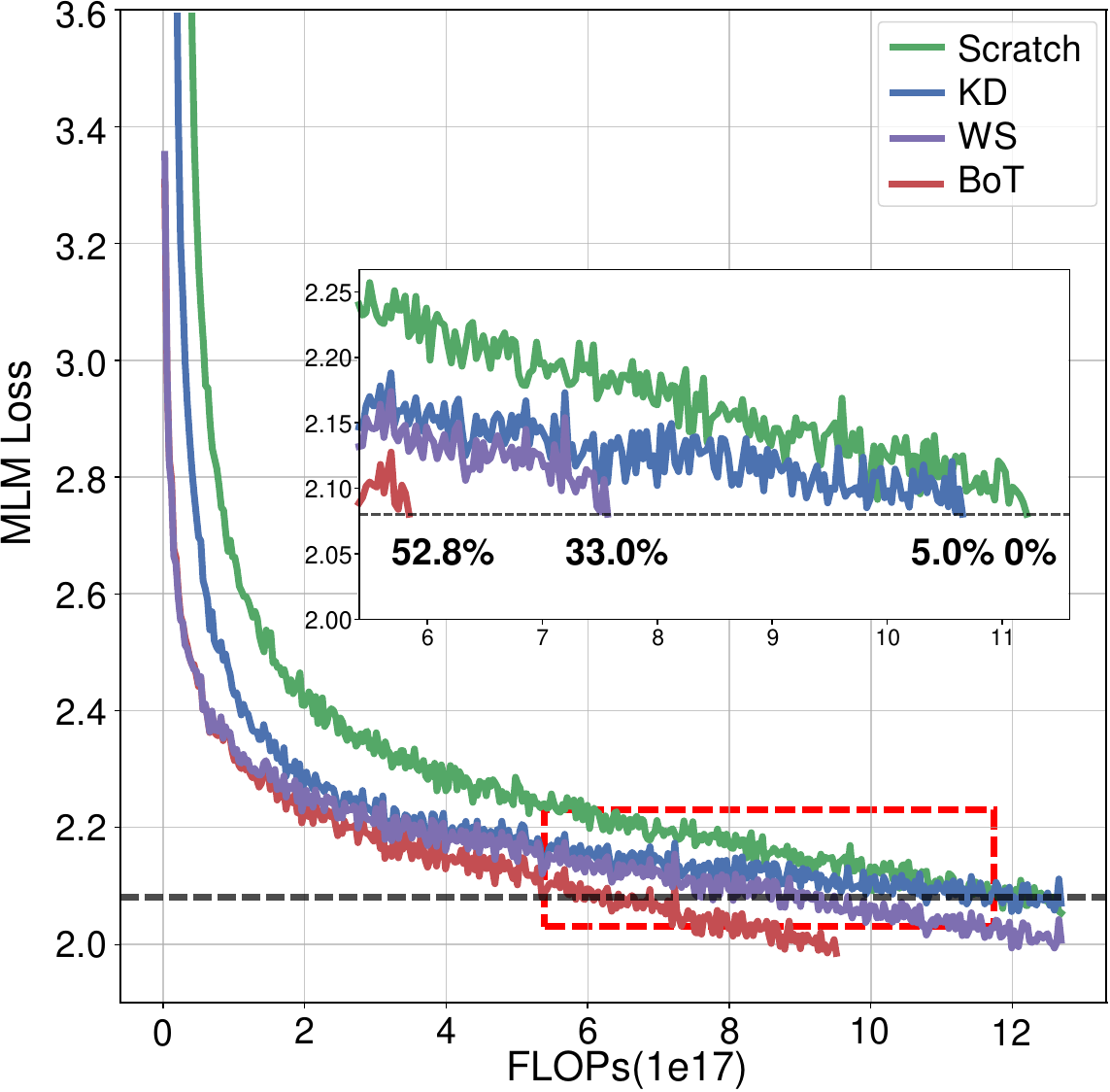}%
    }
    \hfill
    \subfloat[GPT-B $ \rightarrow$ GPT-S\label{fig:gpt_s}]{%
        \includegraphics[width=0.33\textwidth]{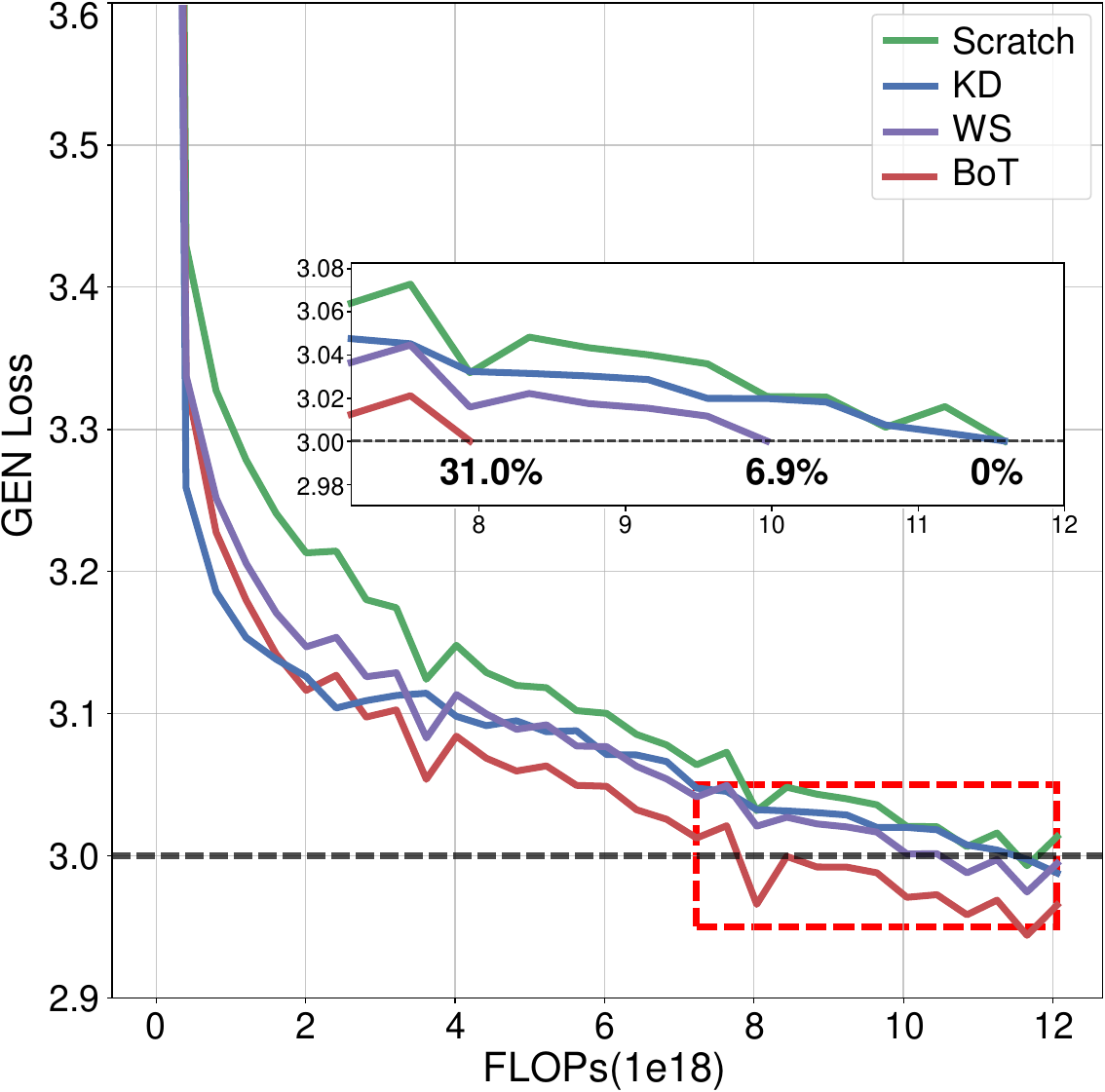}%
    }
    \vspace{-0.07in}
    \caption{Results of pretraining DeiT-S, BERT-S and GPT-S. BoT can achieve the highest savings in FLOPs with 22.0\% for DeiT-S, 67.1\% for BERT-S, and 58.3\% for GPT-S from the Scratch models.}
    \label{fig:small}
    \vspace{-0.17in}
\end{figure*}

Existing Small-to-Large (S2L) methods, shown in Figure~\ref{fig:framework}~(b), often prioritize function preservation. For example, bert2BERT~\cite{chen2021bert2BERT} directly embeds source parameters into the target model. While this provides a strong initial performance, it can lead to distributional discrepancies. The new weights, initialized by splitting or copying existing neurons, may limit the initial state's diversity and constrain the larger model's convergence, as it remains bound to the source model's weight distribution. 
To address this rigidity, trainable methods like LiGO~\cite{wang2023learning} and Mango~\cite{pan2023reusing} learn a projection or transformation to map the source weights to the target space. However, they introduce additional computational overhead and implementation complexity.

In summary, existing L2S approaches are often heuristic, while S2L approaches are either naive or trainable mappers. BoT, in contrast, is proposed as a principled, and training-free framework for addressing both of these challenges.

\textbf{Evaluation Metric.} To quantify computational savings, we introduce the FLOPs Saving Ratio ($r$). This metric compares the FLOPs required to reach a target performance $M$ when using a given method versus training from scratch. This target $M$ represents a specific performance goal. For pre-training language tasks, this could be an objective like Masked Language Modeling (MLM) loss or autoregressive Generation (GEN) loss. For visual tasks, it could be a metric like image classification accuracy.

Let $\xi_{\text{scratch}}$ represent the total FLOPs for a \textit{from-scratch} model to reach $M$, and let $\xi_{*}$ be the FLOPs for an initialized model to reach the same target $M$. The $r$ is calculated as:
\vspace{-0.07in}
\begin{equation*}
    r = \frac{\xi_{\text{scratch}} - \xi_{*}}{\xi_{\text{scratch}}}.
    \vspace{-0.07in}
\end{equation*}
This metric quantifies the fraction of computational cost saved compared to the from-scratch pipeline. A higher $r$ value indicates greater training efficiency.

\subsection{Results on Large-Scale Vision Models}
\label{subsec:vision_model}
To demonstrate training acceleration and bidirectional knowledge transfer on large vision models, we conduct a comprehensive suite of scaling and transferability experiments. Our evaluation is structured by a direct pre-training acceleration benchmark and a downstream fine-tuning evaluation. First, we evaluate both L2S transfer by initializing a DeiT-S with 6 layers from a pre-trained DeiT-B with 12 layers on ImageNet-1K~\cite{krizhevsky2012imagenet}, and S2L transfer by initializing a DeiT-B from a pre-trained DeiT-S. We use Adam as the optimizer with learning rate 2.5e-4 for small and 1.25e-4 for base models and weight decay 5e-2. The batch size is 256 for small and 128 for base models. The training epoch is 150. Second, to assess the quality and transferability of the initialized knowledge, we evaluate the models on a diverse benchmark of 7 downstream datasets. Crucially, we take the models initialized from both the L2S and S2L settings and fine-tune them directly on these tasks, without any intermediate pre-training. This comprehensive benchmark covers a wide spectrum of domains: Oxford Flowers~\citep{nilsback2008automated}, CUB-200-2011~\citep{wah2011caltech}, Stanford Cars~\citep{gebru2017fine}, CIFAR-10, CIFAR-100~\citep{krizhevsky09}, Food-101~\citep{bossard2014food}, and iNaturalist-2019~\citep{tan2019herbarium}. Further details are provided in the Appendix \ref{app:subsec_vision_details}.

\begin{figure*}[tb]
    \centering
    \subfloat[DeiT-S $ \rightarrow$ DeiT-B\label{fig:deit_b}]{%
        \includegraphics[width=0.3275\textwidth]{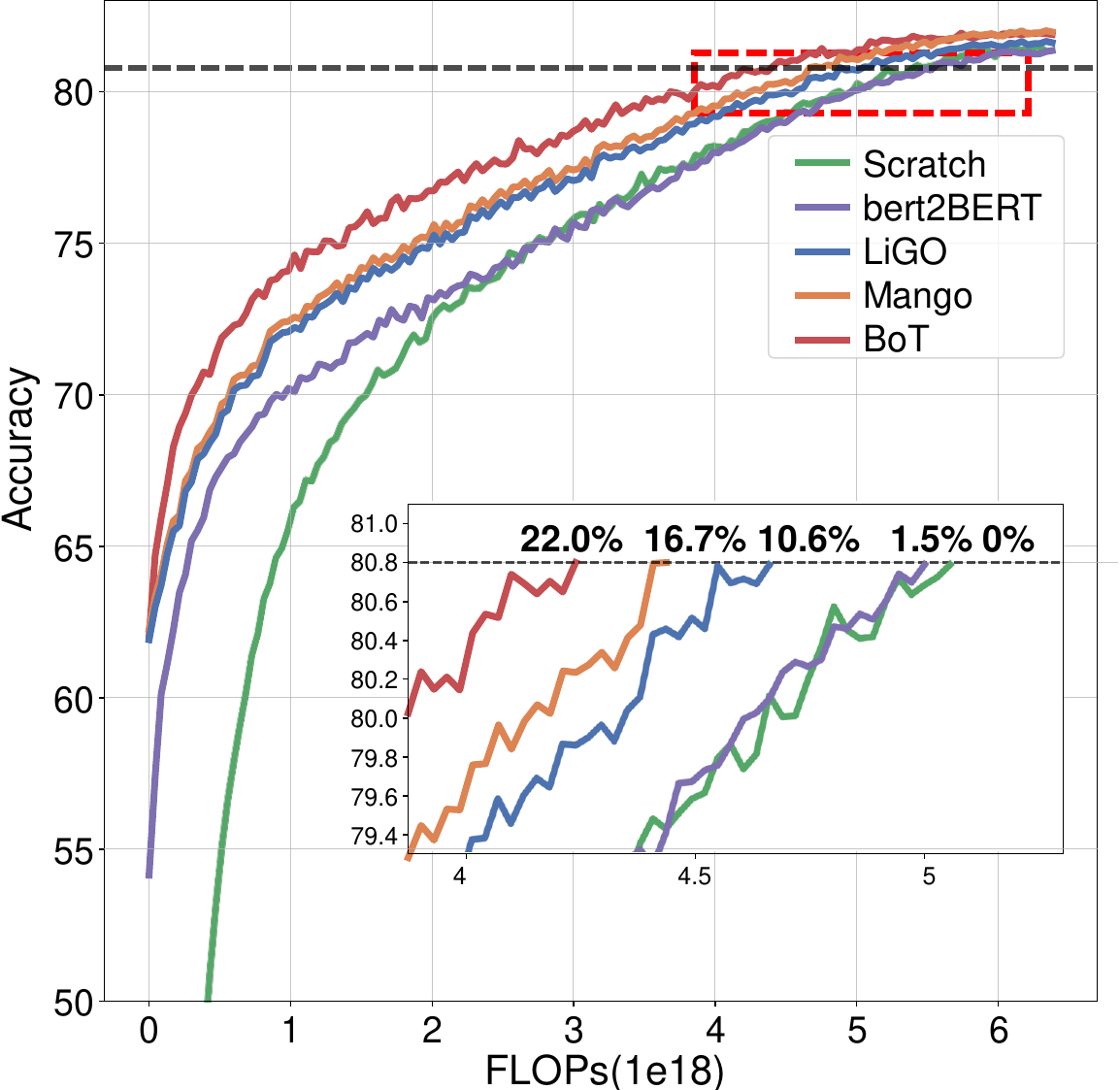}%
    }
    \hfill
    \subfloat[BERT-S $ \rightarrow$ BERT-B\label{fig:bert_b}]{%
        \includegraphics[width=0.33\textwidth]{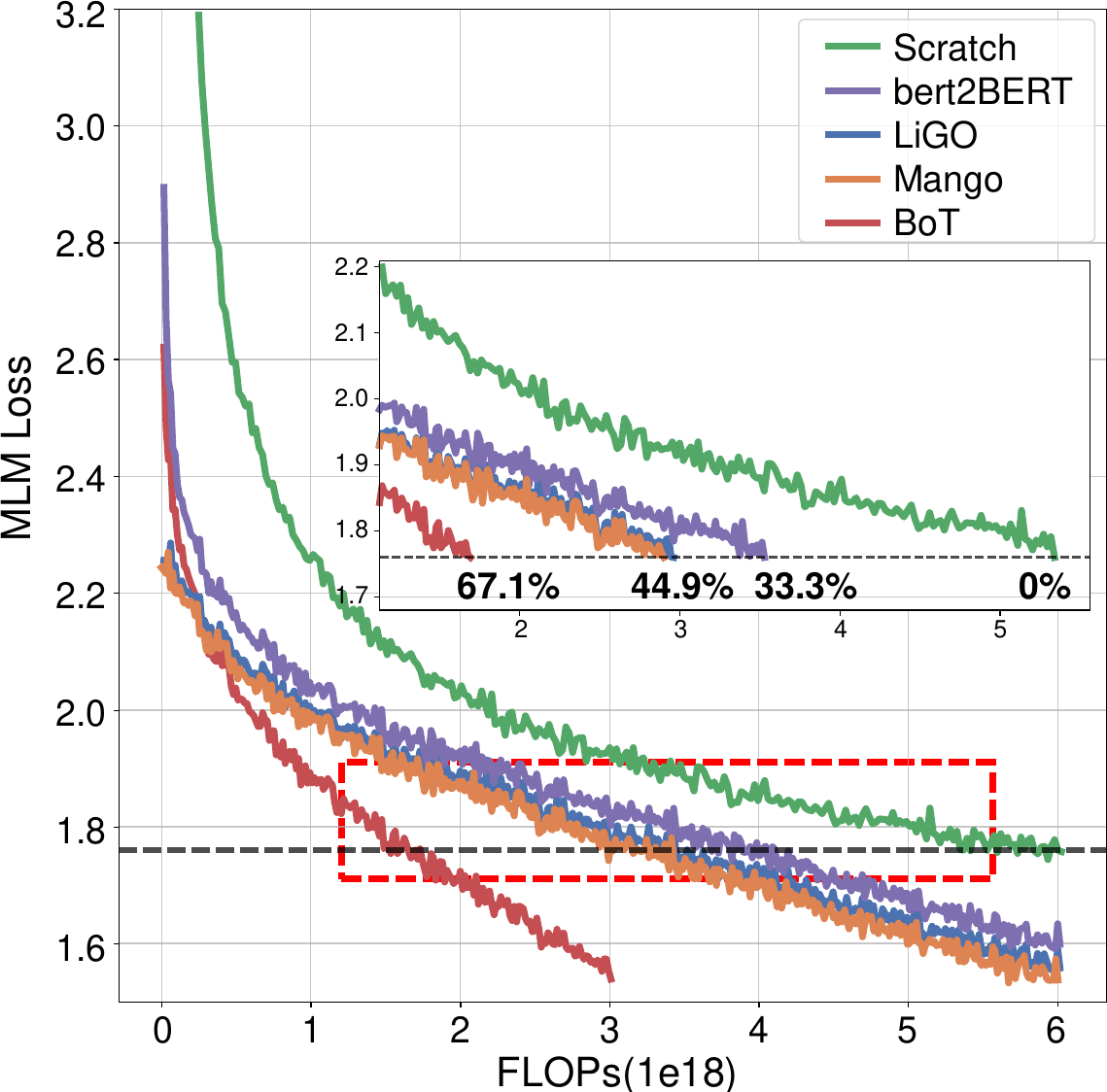}%
    }
    \hfill
    \subfloat[GPT-S $ \rightarrow$ GPT-B\label{fig:gpt_b}]{%
        \includegraphics[width=0.33\textwidth]{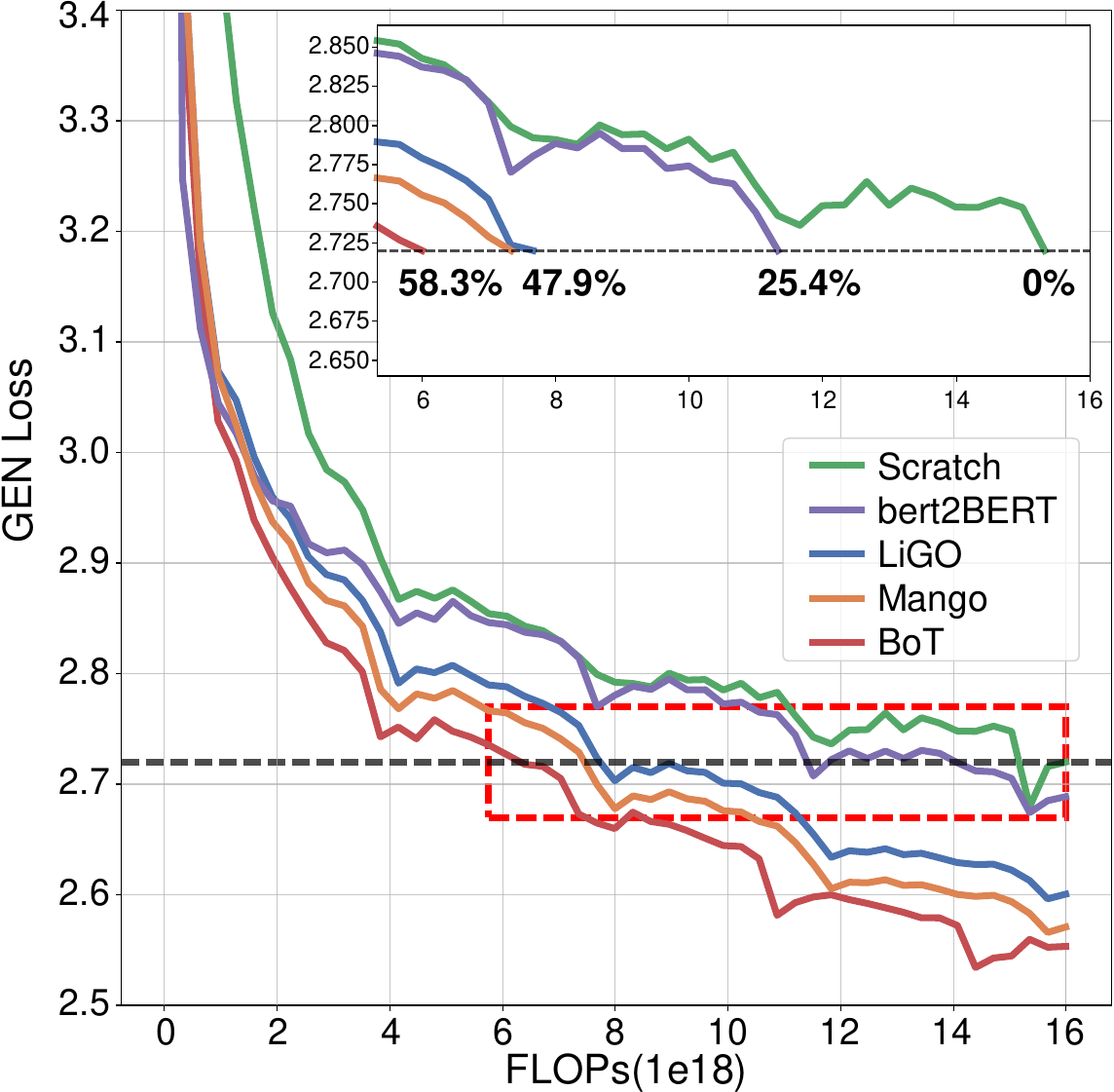}%
    }
    \vspace{-0.07in}
    \caption{Results of pretraining DeiT-B, BRET-B and GPT-B. BoT can achieve the highest savings in FLOPs with 22.0\% for DeiT-B, 67.1\% for BERT-B, and 58.3\% for GPT-B from the Scratch models.}
    \label{fig:base}
    \vspace{-0.17in}
\end{figure*}

As shown in Figure~\ref{fig:deit_s} and~\ref{fig:deit_b}, BoT demonstrates \textbf{superior training efficiency, validating its effectiveness as a unified bidirectional transfer framework}. In the L2S (adaption) scaling, BoT saves a remarkable 39.0\% of FLOPs to reach 70\% accuracy. This significantly surpasses other L2S-specific methods like KD and WS, demonstrating the superiority of our principled decomposition over repetitive teacher guidance and specialized heuristics. Impressively, this strong performance extends to the S2L scaling, where BoT not only achieves the target 80.8\% accuracy with 22.0\% FLOPs savings, but its learning curve, in Figure~\ref{fig:deit_b}, also shows a consistently steeper ascent and higher final convergence. This indicates a more effective initialization, surpassing specialized-S2L methods by a significant margin outperforming LiGO by +11.4\% FLOPs and Mango by +5.3\% FLOPs. This consistently strong performance in both adaptation and expansion validates that our principled, frequency-domain approach effectively transcends the limitations of specialized methods, offering a single, unified framework for arbitrary-sized knowledge transfer.

The results indicate that BoT's superior initialization directly yields \textbf{stronger downstream transferability}, as shown in Table~\ref{tab:downstream_vision}.
In the L2S setting, BoT achieves a 75.1\% average accuracy, dramatically surpassing training from scratch (+18.9\%) and outperforming the WS baseline (+1.2\%). Critically, BoT's advantage is particularly significant on challenging fine-grained tasks, outperforming WS by +3.6\% on CUB and +2.5\% on Cars. This suggests its initialization better preserves the complex, high-resolution features required for these tasks. In the S2L setting, where BoT also attains the highest average accuracy and re-affirms on the most difficult fine-grained datasets, posting a significant margin on CUB and Cars. These results confirm BoT's more effective and transferable initialization, which leverages its unique advantage of preserving structural knowledge critical for complex, fine-grained visual recognition.

\begin{table}
\centering
\setlength{\tabcolsep}{2.6 mm}
\caption{Performance comparison on seven downstream tasks. Models are initialized using various methods and directly fine-tuned without extra pre-training. The ``Avg" column shows the average accuracy.}
\label{tab:downstream_vision}
\vspace{-0.09in}
\resizebox{\linewidth}{!}{
\begin{tabular}{@{}clccccccc@{}}
\toprule[1.5pt]
& Methods & C\small{10} & C\small{100} & CUB & Flow. & Food & iNat. & Cars  \\
\midrule
\multirow{3}{*}{\rotatebox{90}{DeiT-S}}
& Scratch & 94.0 & 66.5 & 27.3 & 57.2 & 70.6 & 54.0 & 23.8  \\
 & WS & 95.8 & 75.2 & 57.7 & 75.8 & 79.2 & 61.2 & 72.1  \\
 &\cellcolor{blue!12}{{BoT}} & \cellcolor{blue!12}\textbf{96.0} & \cellcolor{blue!12}\textbf{75.3} & \cellcolor{blue!12}\textbf{61.3} & \cellcolor{blue!12}\textbf{76.3} & \cellcolor{blue!12}\textbf{80.5} & \cellcolor{blue!12}\textbf{61.9} & \cellcolor{blue!12}\textbf{74.6} \\
\midrule
\multirow{5}{*}{\rotatebox{90}{DeiT-B}}
 & Scratch & 95.9 & 73.8 & 42.2 & 65.0 & 78.6 & 58.2 & 42.1  \\
 & bert2BERT & 96.4 & 74.9 & 67.1 & 85.4 & 82.0 & 61.1 & 82.9  \\
 & LiGO & 97.2 & 82.2 & 73.2 & 94.4 & 84.4 & 67.1 & 87.4  \\
 & Mango & 97.2 & 82.3 & 73.7 & 94.6 & 83.8 & 67.1 & 87.2  \\
 &\cellcolor{blue!12}{BoT} & \cellcolor{blue!12}\textbf{97.2} & \cellcolor{blue!12}\textbf{82.4} & \cellcolor{blue!12}\textbf{74.1} & \cellcolor{blue!12}\textbf{94.7} & \cellcolor{blue!12}\textbf{84.4} & \cellcolor{blue!12}\textbf{67.1} & \cellcolor{blue!12}\textbf{88.6} \\
\bottomrule[1.5pt]
\end{tabular}%
}
\vspace{-0.22in}
\end{table}

\begin{table*}
\centering
\setlength{\tabcolsep}{2.2 mm}
\caption{Performance comparison on the GLUE and SQuAD benchmarks based on BERT. Models are initialized and then directly fine-tuned without additional pre-training. We report accuracy for most tasks, along with Matthews Correlation Coefficient (MCC) for CoLA and Pearson correlation for STS-B. SQuAD performance is measured by F1 scores and Exact Match (EM). The ``Avg'' columns show the average scores for each benchmark.}
\vspace{-0.1in}
\label{tab:downstream_bert}
\resizebox{0.96\textwidth}{!}{
\begin{tabular}{cl ccccccc>{\columncolor{gray!15}}c | cc>{\columncolor{gray!15}}c}
\toprule[1.5pt]
\multirow{2}{*}{} & \multirow{2}{*}{Methods} & \multicolumn{8}{c|}{GLUE Benchmark} & \multicolumn{3}{c}{SQuAD Benchmark} \\
\cmidrule(lr){3-10} \cmidrule(lr){11-13}
& & SST-2 & MNLI & MRPC & CoLA & QNLI & QQP & STS-B & \textbf{Avg.} & v1.1 (F1/EM)  & v2.0 (F1/EM)& \textbf{Avg.} \\
\midrule

\multirow{3}{*}{\rotatebox{90}{{BERT-S}}}
& Scratch & 81.19 & 63.45 & 69.85 & 8.02 & 59.09 & 81.08 & 20.12 & 54.69 & 18.28 / 9.05 & 8.13 / 3.72  & 9.80 \\
& WS & 81.77 & 74.26 & 69.61 & 10.21 & 77.72 & 86.43 & 26.33 & 60.90 & 32.99 / 23.38 &17.11 / 11.65 & 21.28 \\
& \cellcolor{blue!12}{BoT} & \cellcolor{blue!12}\textbf{85.67} & \cellcolor{blue!12}\textbf{74.64} & \cellcolor{blue!12}\textbf{70.34} & \cellcolor{blue!12}\textbf{17.05} & \cellcolor{blue!12}\textbf{81.97} & \cellcolor{blue!12}\textbf{88.13} & \cellcolor{blue!12}\textbf{74.20} & \cellcolor{blue!12}\textbf{70.29} & \cellcolor{blue!12}\textbf{69.63 / 57.87} & \cellcolor{blue!12}\textbf{35.77 / 29.71} & \cellcolor{blue!12}\textbf{48.25} \\
\midrule
\multirow{5}{*}{\rotatebox{90}{{BERT-B}}}
& Scratch & 80.85 & 63.31 & 68.38 & 2.06 & 61.19 & 79.20 & 11.44 & 52.35 &18.95 / 9.62 & 9.23 / 4.55  & 10.59 \\
& bert2BERT & 81.19 & 71.44 & 69.61 & 6.09 & 75.27 & 85.84 & 23.44 & 58.98 &31.58 / 22.20 &   16.25 / 11.20 & 20.31 \\
& LiGO & 85.47 & 77.31 & 73.04 & 25.81 & 84.12 & 88.21 & 75.70 & 72.81 &79.23 / 69.65 & 39.45 / 34.42 & 55.69 \\
& Mango & 85.24 & 77.30 & 73.04 & 25.85 & 84.16 & 88.24 & 75.92 & 72.82 & 79.24 / 69.51 & 39.53 / 34.27 & 55.64 \\
& \cellcolor{blue!12}{BoT} & \cellcolor{blue!12}\textbf{85.78} & \cellcolor{blue!12}\textbf{77.33} & \cellcolor{blue!12}\textbf{74.02} & \cellcolor{blue!12}\textbf{26.34} & \cellcolor{blue!12}\textbf{84.26} & \cellcolor{blue!12}\textbf{88.53} & \cellcolor{blue!12}\textbf{76.40} & \cellcolor{blue!12}\textbf{73.24} & \cellcolor{blue!12}\textbf{80.22 / 70.34} & \cellcolor{blue!12}\textbf{40.99 / 34.50} & \cellcolor{blue!12}\textbf{56.51} \\
\bottomrule[1.5pt]
\end{tabular}%
}
\vspace{-0.17in}
\end{table*}

\subsection{Results on BERT}
\label{subsec:bert}
To validate BoT's effectiveness on encoder-only architectures, we conduct a comprehensive evaluation using the BERT~\cite{devlin2018bert} model. We perform two sets of experiments on the English Wikipedia corpus, focusing on the 12-layer BERT-B and 6-layer BERT-S configurations. First, to evaluate pre-training acceleration, we initialize models for both BERT-B $\to$ BERT-S and BERT-S $\to$ BERT-B transfer. These initialized models are then further pre-trained on the corpus for 400K steps with a batch size of 256. Second, to assess the initialization quality itself, the initialized models are also evaluated by immediately fine-tuning them on the GLUE~\cite{wang2018glue} and SQuAD~\cite{rajpurkar2016squad,rajpurkar2018know} benchmarks. Further Bert details are available in the Appendix~\ref{app:subsec_bert}. In addition, to demonstrate the transferability of our approach, we conduct a systematic parallel evaluation on the RoBERTa model~\cite{liu2019roberta}, as detailed in Appendix~\ref{app:subsec_roberta_results}.

The pre-training acceleration curves, illustrated in Figures~\ref{fig:bert_s} and~\ref{fig:bert_b}, show \textbf{BoT's state-of-the-art bidirectional performance}. 
In the L2S scenario, BoT demonstrates significant acceleration: it reaches the target MLM loss saving 52.8\% FLOPs, outperforming the strong training-free baseline, WS, by +19.8\%. This superiority extends to the S2L direction. BoT achieves the fastest convergence, requiring 67.1\% fewer FLOPs to reach the same performance, a +22.2\% improvement over competing trainable methods like LiGO and Mango. These results underscore our framework's effectiveness. BoT provides a unified solution that drastically reduces the computational cost for the target models to acquire essential language representation capabilities in both transfer directions.

The direct fine-tuning evaluation, presented in Table~\ref{tab:downstream_bert}, confirms that \textbf{BoT provides superior downstream transferability on diverse NLP tasks}. In the L2S scenario on BERT-S, BoT dramatically outperforms the WS baseline. It achieves a substantial +9.39\% lead on the GLUE average and, notably, more than doubles the SQuAD average score for a gain of +26.97\%. This strong performance extends to the S2L direction (BERT-B), where BoT again secures the highest scores on both GLUE (73.24\%) and SQuAD (56.51\%), surpassing all competing trainable expansion methods, including LiGO and Mango. By successfully transferring coherent structural knowledge, BoT-initialized models demonstrate significantly stronger capabilities on complex downstream tasks, ranging from natural language understanding to question answering.

\subsection{Results on GPT2}
\label{subsec:gpt2}

To further assess BoT's efficiency beyond encoder-based models, we extend our evaluation to the GPT-2 architecture~\cite{radford2019language}, a decoder-only auto-regressive model. We conduct bidirectional pre-training experiments on the dataset combining the English Wikipedia and the Toronto Book Corpus~\cite{pan2023reusing}, encompassing both L2S and S2L transfers between GPT-S with 6 layers and GPT-B with 12 layers. More details are shown in the Appendix~\ref{app:subsec_gpt2}.

\textbf{BoT's strong training acceleration extends to GPT-2 models}, presented in Figure~\ref{fig:gpt_s} and Figure~\ref{fig:gpt_b}. 
In the L2S scenario, BoT achieves a 31.0\% FLOPs saving, outperforming the WS baseline by a significant +24.1\%. This strong performance extends to the S2L direction, where our method achieves the highest training acceleration with a 58.3\% FLOPs saving, representing a +10.4\% improvement over competing expansion methods like LiGO and Mango. Notably, BoT consistently maintains the lowest loss throughout training, despite the fundamental architectural differences between GPT's decoder-only stack and BERT's encoder. This significant and consistent performance across disparate architectures strongly validates BoT's robustness, establishing it as a truly unified, architecture-agnostic solution for cross-size initialization.


\subsection{Ablation and Study}
\label{subsec:ablation_study}

\subsubsection{Analysis of Different Wavelet Families}
\label{subsubsec:ablation_study}

\begin{table*}[t]
\centering
\setlength{\tabcolsep}{1.2 mm}
\caption{Performance comparison on GLUE and SQuAD benchmarks after pre-training from initialization. This table shows the final performance of models that were first pre-trained and then fine-tuned. ``Avg.'' columns show the average scores for each benchmark.}
\vspace{-0.1in}
\label{tab:post_pretraining_results}

\resizebox{0.96\textwidth}{!}{
\begin{tabular}{@{}cl cc | ccccccc>{\columncolor{gray!15}}c | cc>{\columncolor{gray!15}}c@{}}
\toprule[1.5pt]
\multirow{2}{*}{} & \multirow{2}{*}{Methods} & \multicolumn{2}{c|}{Savings} & \multicolumn{8}{c|}{GLUE Benchmark} & \multicolumn{3}{c}{SQuAD Benchmark} \\
\cmidrule(lr){3-4} \cmidrule(lr){5-12} \cmidrule(lr){13-15}
& & FLOPs & Walltime & SST-2 & MNLI & MRPC & CoLA & QNLI & QQP & STS-B & \textbf{Avg.} & v1.1 (F1/EM) & v2.0 (F1/EM) & \textbf{Avg.} \\
\midrule

\multirow{3}{*}{\rotatebox{90}{{BERT-S}}}
& Scratch & 0.0\% & 0.0\% & 87.50 & 77.74 & 77.94 & 28.47 & 85.54 & 88.93 & 83.01 & 75.59 & 76.12 / 66.02 & 37.86 / 32.43 & 53.11 \\
& WS & 33.0\% & 41.7\% & 89.45 & 77.67 & 77.70 & 28.00 & 85.14 & 88.93 & 82.15 & 75.58 & 78.07 / 68.41 & 39.43 / 34.27 & 55.04 \\
& \cellcolor{blue!12}{BoT} & \cellcolor{blue!12}\textbf{52.8\%} & \cellcolor{blue!12}\textbf{66.8\%} & \cellcolor{blue!12}\textbf{89.33} & \cellcolor{blue!12}\textbf{78.34} & \cellcolor{blue!12}\textbf{77.47} & \cellcolor{blue!12}\textbf{27.54} & \cellcolor{blue!12}\textbf{85.69} & \cellcolor{blue!12}\textbf{88.95} & \cellcolor{blue!12}\textbf{82.62} & \cellcolor{blue!12}\textbf{75.71} & \cellcolor{blue!12}\textbf{78.79 / 68.87} & \cellcolor{blue!12}\textbf{39.52 / 34.23} & \cellcolor{blue!12}\textbf{55.35} \\
\midrule

\multirow{5}{*}{\rotatebox{90}{{BERT-B}}}
& Scratch & 0\% & 0.0\% & 91.51 & 83.28 & 84.31 & 43.34 & 89.93 & 90.64 & 85.47 & 81.21 & 84.81 / 75.33 & 45.43 / 40.81 & 61.59 \\
& bert2BERT & 33.3\% & 32.4\% & 90.48 & 83.60 & 84.80 & 50.47 & 90.19 & 90.44 & 87.03 & 82.43 & 84.83 / 75.86 & 45.51 / 41.07 & 61.82 \\
& LiGO & 44.7\% & 43.5\% & 90.60 & 83.09 & 84.56 & 45.42 & 90.12 & 90.55 & 86.83 & 81.60 & 85.48 / 76.32 & 42.59 / 37.34 & 60.41 \\
& Mango & 44.9\% & 43.7\% & 90.25 & 83.39 & 84.07 & 45.35 & 89.82 & 90.64 & 87.12 & 81.52 & 85.27 / 76.52 & 42.66 / 37.49 & 60.49 \\
& \cellcolor{blue!12}{BoT} & \cellcolor{blue!12}\textbf{67.1\%} & \cellcolor{blue!12}\textbf{65.3\%} & \cellcolor{blue!12}\textbf{90.94} & \cellcolor{blue!12}\textbf{83.46} & \cellcolor{blue!12}\textbf{87.99} & \cellcolor{blue!12}\textbf{48.12} & \cellcolor{blue!12}\textbf{90.06} & \cellcolor{blue!12}\textbf{90.66} & \cellcolor{blue!12}\textbf{87.70} & \cellcolor{blue!12}\textbf{82.70} & \cellcolor{blue!12}\textbf{85.02 / 75.84} & \cellcolor{blue!12}\textbf{45.17 / 40.39} & \cellcolor{blue!12}\textbf{61.61} \\
\bottomrule[1.5pt]
\end{tabular}%
}
\vspace{-0.17in}
\end{table*}

We conducted an ablation study to investigate how the choice of wavelet type impacts our initialization's effectiveness. We performed bidirectional transfers on both an encoder model (BERT) and a decoder model (GPT), measuring validation loss after a fixed number of pre-training steps. More details are shown in Appendix~\ref{app:sec_wavelet_family}.

\begin{figure}[tb]
    \centering

    \includegraphics[width=\columnwidth]{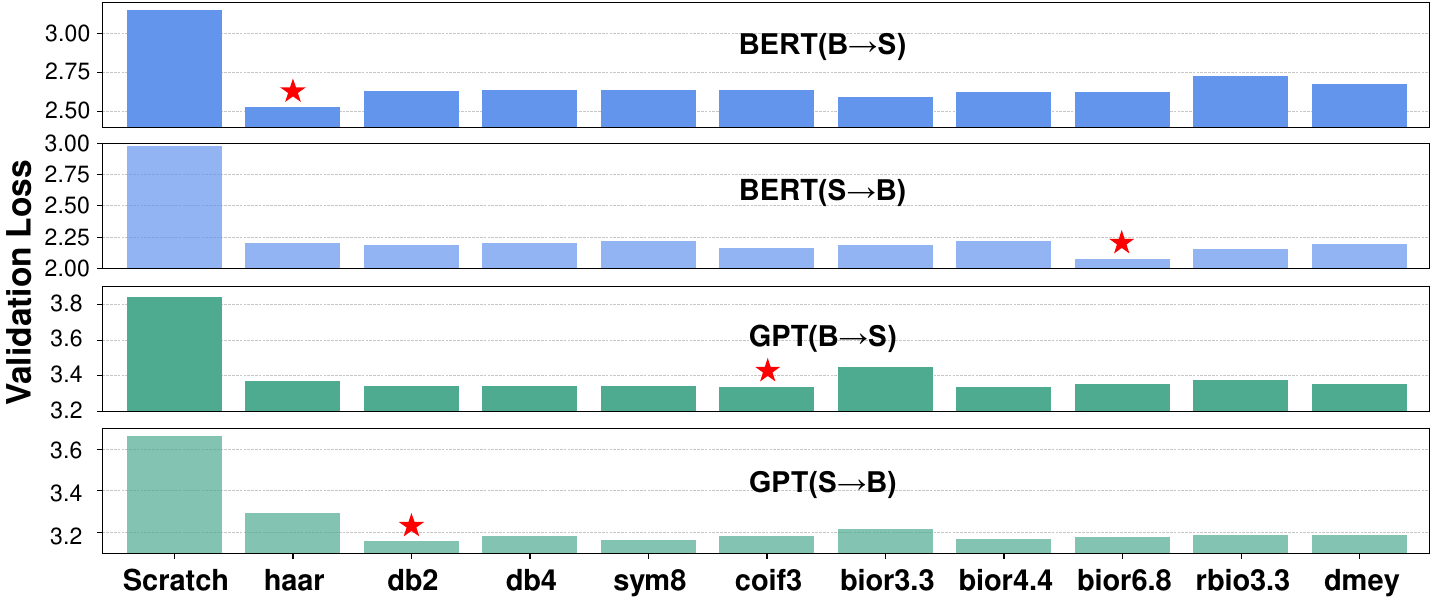}
    \caption{Ablation study on the choice of wavelet family. Lower validation loss indicates better performance. }
    \label{fig:wavelet}
    \vspace{-0.27in}

\end{figure}

As shown in Figure~\ref{fig:wavelet}, the results reveal that the \textit{optimal} wavelet choice is context-specific, depending on both architecture and transfer direction. For the BERT models, performance is strongly tied to wavelet properties. In the L2S setting, the simplest wavelet, Haar, is optimal, suggesting its piecewise-constant nature is highly effective in encoders. For L2S  transfer, wavelets with higher-order vanishing moments and greater smoothness show a clear advantage, with Biorthogonal (bior6.8) emerging as the top performer. In contrast, the auto-regressive decoder model (GPT) generally benefits from wavelets with compact support. The Coiflet (coif3)  wavelet proves optimal for its L2S transfer, while Daubechies (db2) excels in the S2L direction. These findings highlight that encoder expansion uniquely favors smoother, higher-order wavelets, while other transfers generally perform better with simpler, shorter-filter wavelets.

A secondary, but fundamental, observation is the overall robustness of the wavelet-based approach itself. In all four tested configurations, the ``Scratch" baseline consistently yielded the highest validation loss, whereas every wavelet-initialized models achieved significantly better performance. This strongly underscores the fundamental robustness and efficacy of the wavelet-based approach itself.

\subsubsection{Transferability with Pretraining on BERT}
\label{subsec:transferability_with_pretraining}

To evaluate BoT on final task performance after pretraining, we first pre-train to the same MLM loss as the initialized models and then fine-tune them on downstream tasks. The final performance, alongside the computational cost savings during pre-training, is presented in Table~\ref{tab:post_pretraining_results}. The ``Walltime" metric measures the actual clock time saved, which, unlike FLOPs, also accounts for system-level efficiencies.

In both L2S on BERT-S and S2L on BERT-B scenarios, BoT achieves substantial efficiency gains. For BERT-S, BoT saves 52.8\% of FLOPs and a remarkable 66.8\% of Walltime, significantly outperforming the WS baseline while achieving a state-of-the-art average score of 75.71\% on GLUE and 55.35\% on SQuAD. Similarly, for BERT-B, BoT leads with 67.1\% FLOPs and 65.3\% Walltime savings, far exceeding all other expansion methods. Critically, despite this massive reduction in computational cost, the final downstream performance of BoT remains on par with, or even slightly better than, more training needed methods like LiGO and Mango. This demonstrates that BoT achieves equivalent or superior transfer performance while converging significantly faster, offering a highly practical and resource-efficient solution for knowledge transfer.

\subsection{Visualization of Knowledge}
\label{subsec:visualization}



\begin{figure}[tb]
    \centering
    \includegraphics[width=0.98\columnwidth]{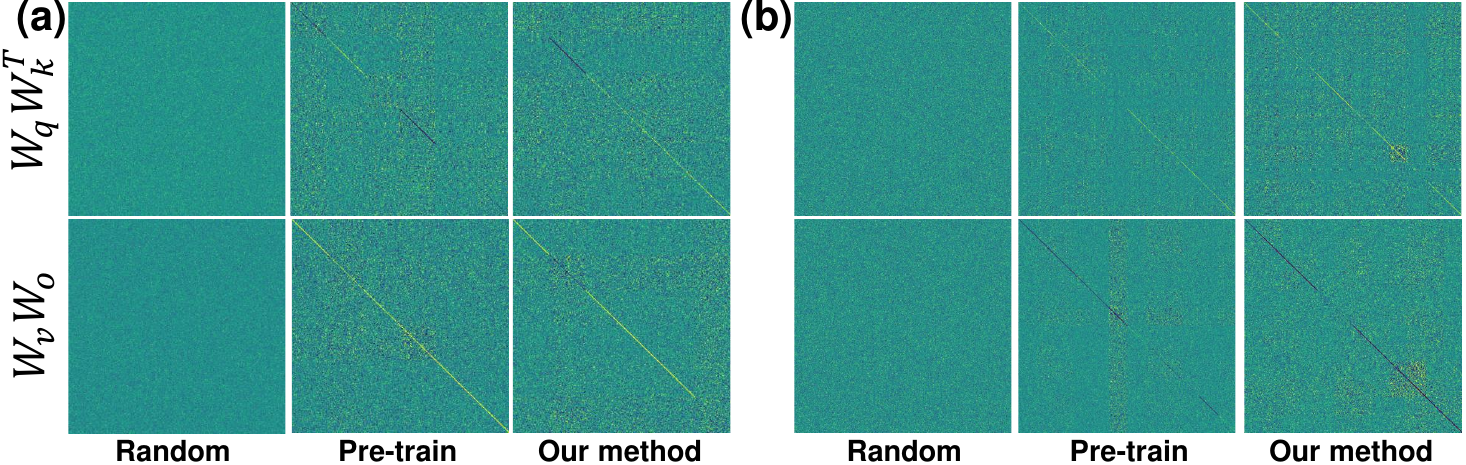}
    \vspace{-0.1in}
    \caption{Visualization of structured knowledge within self-attention layers for (a) DeiT-S and (b) DeiT-B. BoT inherits the strong diagonal structure present in pre-trained models.}
    \label{fig:visual}
    \vspace{-0.22in}
\end{figure}

As illustrated in Figure~\ref{fig:visual}, BoT enables models to inherit the essential diagonal attributes within self-attention layers, a characteristic typically exclusive to pre-trained models. BoT autonomously encapsulates structured knowledge from pre-trained models into the learngene without requiring manual intervention.


\begin{figure}[tb]
    \centering
    \includegraphics[width=0.96\columnwidth]{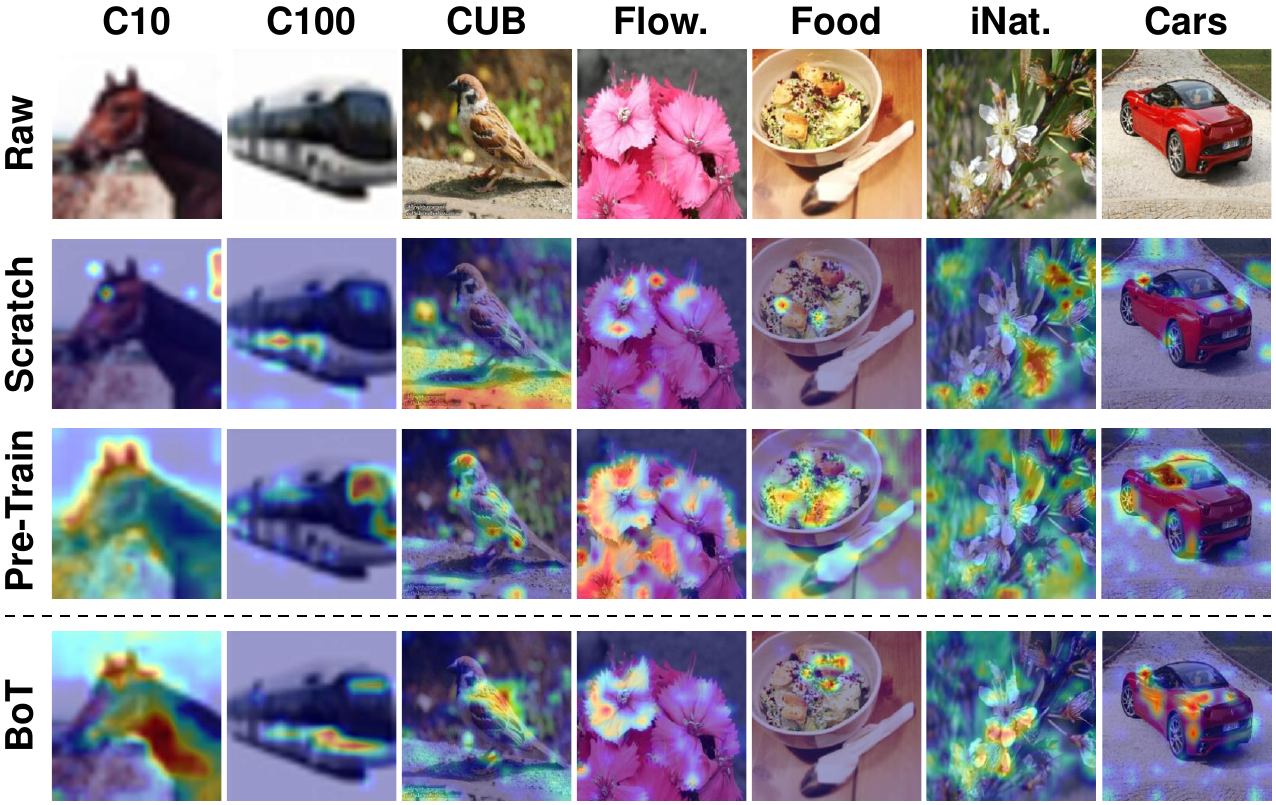}
    \vspace{-0.12in}
    \caption{Visualization of knowledge extracted by BoT. All networks operate directly after initialization without any additional training or fine-tuning.}
    \label{fig:cam_visualization}
    \vspace{-0.22in}

\end{figure}

Furthermore, we demonstrate that BoT encourages the initialized model to concentrate on salient local features. We employ Class Activation Mapping (CAM)~\cite{selvaraju2017grad} to visualize the model's attention on sample images, comparing the focus of a model initialized with BoT against both a randomly initialized (Scratch) model and a fully pre-trained one. As illustrated in Figure~\ref{fig:cam_visualization}, the Scratch model exhibits scattered, unfocused attention across the image. The pre-trained model, conversely, transfers a broad set of features, resulting in wide attention that often incorrectly includes irrelevant background details. In contrast, the BoT-initialized model demonstrates superior localization, with its attention concentrated in smaller, more precise regions corresponding to the object of interest. This cleaner focus, which effectively ignores the background, highlights BoT's ability to transfer core, object-centric knowledge, thereby enhancing downstream classification performance.

\section{Conclusion}
\label{sec:conclusion}


In this paper, we addressed the challenge of scaling initialization, where Large-to-Small (L2S) and Small-to-Large (S2L) knowledge transfer are typically treated as separate, incompatible problems.
We proposed BoT, a unified framework that treats model weights as continuous signals, with models of different sizes acting as multi-resolution discretizations of this intrinsic knowledge. This perspective naturally led us to the DWT/IDWT as a single, principled, and parameter-free mechanism for both L2S downsampling and S2L upsampling. BoT demonstrates optimal transferring ability, resulting in state-of-the-art performance in both training acceleration and downstream tasks.

\section*{Acknowledge}
\label{sec:acknowledge}

This research was supported by the Jiangsu Science Foundation (BG2024036, BK20243012, BK20241297), the National Science Foundation of China (62125602, 62406066, U24A20324, 92464301, 625B2045), the New Cornerstone Science Foundation through the XPLORER PRIZE, and the Fundamental Research Funds for the Central Universities (2242025K30024).

{
    \small
    \bibliographystyle{ieeenat_fullname}
    \bibliography{main}
}

\clearpage
\appendix
\setcounter{page}{1}
\maketitlesupplementary

\section{The Discrete Wavelet Transform}
\label{app:dwt}

This appendix provides a detailed mathematical background for the Discrete Wavelet Transform (DWT) and its inverse (IDWT), building from the 1D case to the 3D case used in our work.

\subsection{The 1D Discrete Wavelet Transform}

The 1D-DWT is the fundamental building block. It decomposes a discrete signal into two components: a low-frequency approximation and a high-frequency detail. This is achieved through filtering and down-sampling.

\subsubsection{Decomposition (Analysis)}
Given a discrete signal $x[n]$ of length $N$, the decomposition involves two main steps:

\begin{enumerate}
    \item \textbf{Filtering:} The signal is passed through two complementary filters: a low-pass filter $g[n]$ (associated with the scaling function) and a high-pass filter $h[n]$ (associated with the wavelet function). This is performed via convolution:
    \begin{align*}
        y_{\text{low}}[n] &= (x * g)[n] = \sum_{k} x[k] \cdot g[n-k] \\
        y_{\text{high}}[n] &= (x * h)[n] = \sum_{k} x[k] \cdot h[n-k]
    \end{align*}

    \item \textbf{Down-sampling:} The output of each filter contains redundant information. To create a compact representation, both filtered signals are down-sampled by a factor of $f$ (denoted by $\downarrow f$), meaning only every other sample is kept.
\end{enumerate}

This process yields the \textbf{approximation coefficients} (${cA}$) from the low-pass filter and the \textbf{detail coefficients} (${cD}$) from the high-pass filter. Each set of coefficients has a length of approximately $N/2$. The complete formulas are:
\begin{equation}
\label{eq:appendix_dwt_1d}
\begin{aligned}
    {cA}[k] &= (x * g)_{\downarrow f}[k] = \sum_{n} x[n] \cdot g[2k - n] \\
    {cD}[k] &= (x * h)_{\downarrow f}[k] = \sum_{n} x[n] \cdot h[2k - n]
\end{aligned}
\end{equation}

\subsubsection{Reconstruction (Synthesis)}
The Inverse DWT (IDWT) perfectly reconstructs the original signal from its approximation and detail coefficients. This process reverses the decomposition steps.

\begin{enumerate}
    \item \textbf{Up-sampling:} First, both the ${cA}$ and ${cD}$ coefficient vectors are up-sampled by a factor of f (denoted by $\uparrow f$). This is achieved by inserting a zero between every sample, restoring them to the original signal's length $N$.

    \item \textbf{Filtering:} The up-sampled signals are then passed through corresponding synthesis filters, $g'[n]$ (low-pass) and $h'[n]$ (high-pass).
\end{enumerate}

The outputs from these two filters are summed to perfectly reconstruct the original signal, $x[n]$. The complete reconstruction formula for the signal $x'[n]$ is:
\begin{equation}
\label{eq:appendix_idwt_1d}
\begin{aligned}
    x'[n] &= ({cA}_{\uparrow f} * g')[n] + ({cD}_{\uparrow f} * h')[n] \\
           &= \sum_{k} {cA}[k] \cdot g'[n-2k] + \sum_{k} {cD}[k] \cdot h'[n-2k]
\end{aligned}
\end{equation}

\subsection{Extension to Multi-Dimensional DWT}
\label{subsec:appendix_multi_d}

The multi-dimensional DWT is implemented by applying the 1D-DWT \textbf{separably} along each dimension of the input.

\paragraph{2D-DWT} To illustrate the separable approach, consider a 2D input (image) ${T}$.
\begin{enumerate}
    \item First, the 1D-DWT is applied to each \textbf{row} of ${T}$. This results in two intermediate output: one containing the approximation coefficients of the rows (${cA}_{\text{rows}}$) and one containing the detail coefficients (${cD}_{\text{rows}}$).
    \item Next, the 1D-DWT is applied to each \textbf{column} of these two intermediate weights.
\end{enumerate}
Applying the 1D-DWT to the columns of ${cA}_{\text{rows}}$ yields the final approximation coefficients (${cA} = {T}_{LL}$) and the vertical detail coefficients (${cD}_{\text{vertical}} = {T}_{LH}$). Applying it to the columns of ${cD}_{\text{rows}}$ yields the horizontal details (${cD}_{\text{horizontal}} = {T}_{HL}$) and the diagonal details (${cD}_{\text{diagonal}} = {T}_{HH}$). Thus, a 2D input is decomposed into one approximation sub-band (${cA}$) and three detail sub-bands (${cD}$).

\paragraph{3D-DWT} The 3D-DWT extends this principle to one more dimension. For a 3D input ${T}$:
\begin{enumerate}
    \item Apply the 1D-DWT along the first dimension (\eg, rows, axis $i$).
    \item Apply the 1D-DWT along the second dimension (\eg, columns, axis $j$) to the results of the first step.
    \item Apply the 1D-DWT along the third dimension (\eg, depth, axis $k$) to the results of the second step.
\end{enumerate}
This three-stage separable process results in one low-frequency approximation sub-band (${T}_{LLL}$) and seven high-frequency detail sub-bands (${T}_{LLH}, {T}_{LHL}, {T}_{LHH}, \dots, {T}_{HHH}$).

\subsection{Connection to Main Paper's Notation}
\label{subsec:appendix_notation_link}

The operator notation used in Section \ref{subsec:preliminaries} of the main paper provides a compact representation of this separable filtering process.
\begin{itemize}
    \item The operator $\Phi_d$ represents the application of the 1D-DWT's low-pass analysis stage (filtering with $g$ and down-sampling by f) along dimension $d$.
    \item The operator $\Psi_d$ represents the application of the 1D-DWT's high-pass analysis stage (filtering with $h$ and down-sampling by f) along dimension $d$.
\end{itemize}
Therefore, Equation~(\ref{eq:dwt_3d_approx}) from the main paper,
\begin{equation*}
    {cA} = \Phi_k(\Phi_j(\Phi_i({T})))
\end{equation*}
is the formal mathematical notation for sequentially applying the low-pass filter transform along the $i$, then $j$, and finally $k$ axes of the input ${T}$. Similarly, the various detail coefficients (${cD}_m$) are derived from applying different combinations of $\Phi_d$ and $\Psi_d$ operators. The IDWT operation reverses this process stage by stage, as described conceptually in Section \ref{subsec:appendix_multi_d}.

\section{Overview of Wavelet Families}
\label{app:sec_wavelet_family}

\begin{table}[t]
\centering
\caption{Overview of Wavelet Families Used in Ablation Study. The "Visualization" column shows the wavelet function ($\psi$), which determines how high-frequency details are captured.}
\label{tab:wavelet_visualization}

\resizebox{0.48\textwidth}{!}{
\begin{tabular}{@{}ccm{5cm}c@{}}
\toprule
\textbf{Wavelet} & \textbf{Length} & \textbf{Key Characteristics} & \textbf{Visualization} \\
\midrule

haar & 2 & Discontinuous, piecewise-constant. Has the most compact support. Highly efficient. \newline Optimal for: BERT (L2S) & 
\includegraphics[height=1.5cm]{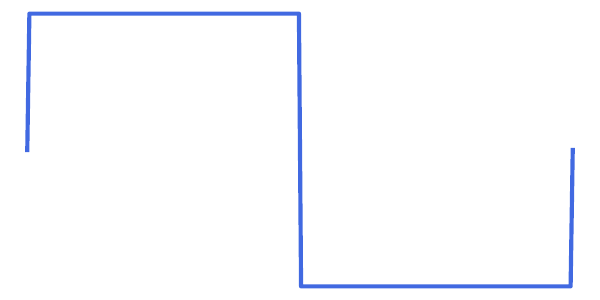} \\
\midrule

db2 & 4 & Daubechies family. Orthogonal and compact, but asymmetric. \newline Optimal for: GPT (S2L) & 
\includegraphics[height=1.5cm]{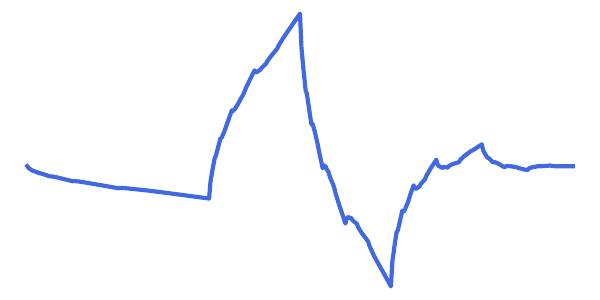} \\
\midrule

db4 & 8 & Daubechies family. Smoother than db2 due to longer filter and more vanishing moments. & 
\includegraphics[height=1.5cm]{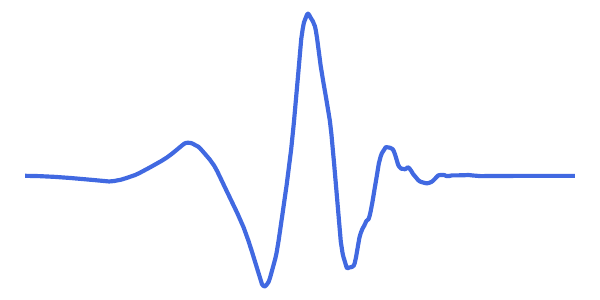} \\
\midrule

sym8 & 16 & Symlet family. ``Least asymmetric" modification of Daubechies wavelets. Orthogonal with compact support. & 
\includegraphics[height=1.5cm]{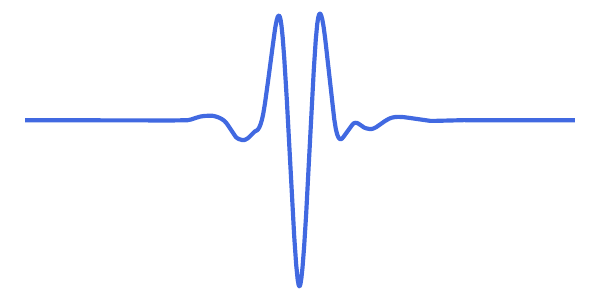} \\
\midrule

coif3 & 18 & Coiflet family. Nearly symmetric with vanishing moments for both scaling and wavelet functions. \newline Optimal for: GPT (L2S) & 
\includegraphics[height=1.5cm]{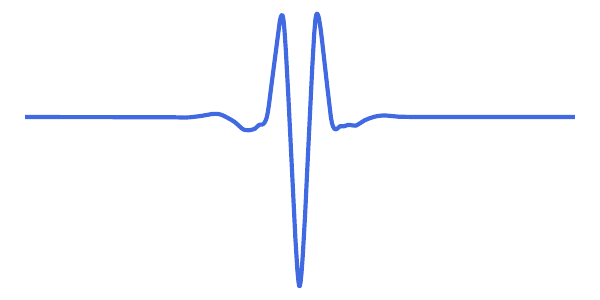} \\
\midrule

bior3.3 & 8 (Decomp.) & Biorthogonal family. Symmetric (linear phase) by relaxing orthogonality. Uses separate analysis/synthesis filters. & 
\includegraphics[height=1.5cm]{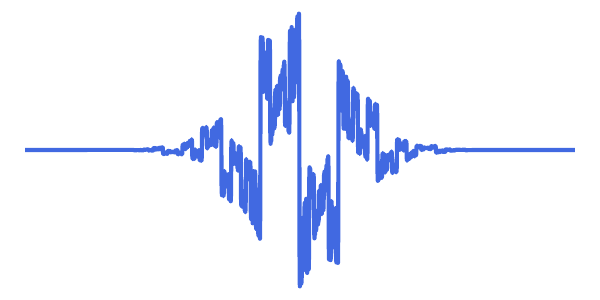} \\
\midrule

bior4.4 & 10 (Decomp.) & Biorthogonal family. Smoother and longer filter than bior3.3. & 
\includegraphics[height=1.5cm]{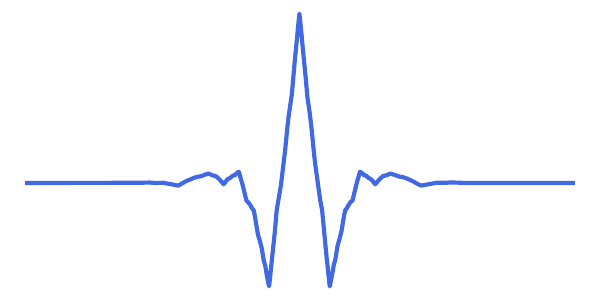} \\
\midrule

bior6.8 & 18 (Decomp.) & Biorthogonal family. Highly smooth with a long filter, ideal for interpolation. \newline Optimal for: BERT (S2L) & 
\includegraphics[height=1.5cm]{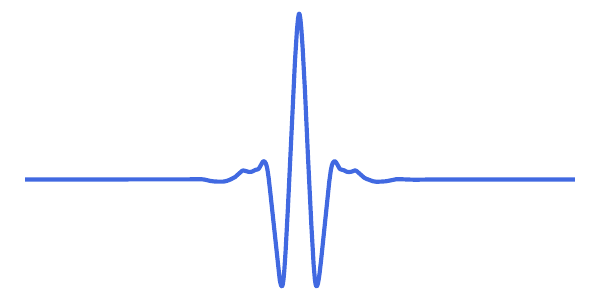} \\
\midrule

rbio3.3 & 8 (Decomp.) & Reverse Biorthogonal. Swaps the decomposition and reconstruction filters of bior3.3. & 
\includegraphics[height=1.5cm]{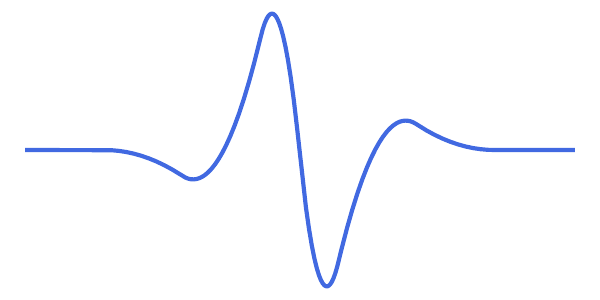} \\
\midrule

dmey & (Approximated) & Discrete Meyer. FIR approximation of an infinitely smooth, orthogonal wavelet defined in the frequency domain. & 
\includegraphics[height=1.5cm]{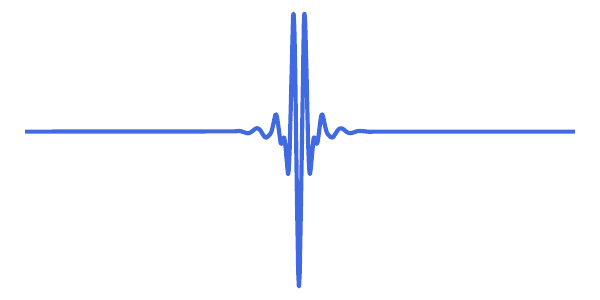} \\

\bottomrule
\end{tabular}
}
\end{table}

In our ablation study (Section~\ref{subsubsec:ablation_study}), we evaluated several common wavelet families provided by the PyWavelets library~\cite{lee2019pywavelets}, as shown in Table~\ref{tab:wavelet_visualization}. The choice of wavelet family is critical as it determines the properties of the low-pass and high-pass filters used in the DWT and IDWT, such as their length, support, symmetry, and smoothness. These properties, in turn, influence how information is decomposed and reconstructed. Below, we provide a brief overview of the families used in our experiments.

\paragraph{Haar Wavelet.}
The Haar wavelet is the simplest and first known wavelet. It is a discontinuous, piecewise-constant function that resembles a step function.
\begin{itemize}
\item \textbf{Properties:} It has the most compact support possible (length 2 filter) and is the only real, orthogonal, and symmetric wavelet (excluding trivial examples). Its primary drawback is its lack of continuity, which can introduce blocky artifacts in signal processing applications.
\item \textbf{Relevance to Our Work:} Its simplicity and compact support make it extremely computationally efficient. Our results suggest its piecewise-constant nature is surprisingly effective for compressing the knowledge in encoder-based models like BERT (L2S).
\end{itemize}

\paragraph{Daubechies Family (db).}
Named after their creator, Ingrid Daubechies, this family is a cornerstone of modern wavelet theory~\cite{daubechies1992wavelets}. The ``dbN" wavelets (\eg, ``db2", ``db4") are orthogonal wavelets where ``N" refers to the number of vanishing moments, which is half the filter length.
\begin{itemize}
\item \textbf{Properties:} They are characterized by compact support and increasing smoothness as ``N" increases. However, they are asymmetric (except for Haar, which is ``db1"), which can sometimes be a disadvantage.
\item \textbf{Relevance to Our Work:} The `db` family offers a direct way to trade off between filter length and smoothness. Our experiments show that the compact ``db2" is particularly effective for expanding decoder-based models like GPT (S2L).
\end{itemize}

\paragraph{Symlets (sym) and Coiflets (coif).}
The Symlet and Coiflet families were designed to be modifications of the Daubechies wavelets to improve symmetry.
\begin{itemize}
\item \textbf{Properties:} ``symN" wavelets are ``least asymmetric" for their support length, making them a popular choice when near-symmetry is desired. ``coifN" wavelets are constructed to have both the scaling function and wavelet function possess vanishing moments, which can be useful for compression.
\item \textbf{Relevance to Our Work:} ``sym8" and ``coif3" represent wavelets with longer filters and greater smoothness compared to ``db2"/``db4". Our results show them to be strong performers in several settings, particularly ``coif3" for GPT (L2S).
\end{itemize}

\paragraph{Biorthogonal Family (bior) and Reverse Biorthogonal (rbio).}
This family relaxes the orthogonality constraint, instead using two different sets of wavelets for decomposition and reconstruction (analysis and synthesis). This trade-off allows for the construction of wavelets that are both symmetric and have compact support (unlike the ``db" family).
\begin{itemize}
\item \textbf{Properties:} The key advantage is linear phase, which means they do not introduce phase distortion. The notation ``biorNr.Nd" indicates the filter lengths for reconstruction (``Nr") and decomposition (``Nd"). The ``rbio" family is simply the reverse pairing.
\item \textbf{Relevance to Our Work:} The flexibility of this family is evident in our results. The highly smooth ``bior6.8" proves to be the best for expanding encoder-based models (BERT S2L), suggesting its longer, smoother filters are ideal for interpolating complex, bidirectional knowledge.
\end{itemize}

\paragraph{Discrete Meyer Wavelet (dmey).}
The Meyer wavelet is an infinitely smooth, orthogonal wavelet defined in the frequency domain. ``dmey" is its FIR (Finite Impulse Response) approximation.
\begin{itemize}
\item \textbf{Properties:} It is known for its excellent smoothness properties.
\item \textbf{Relevance to Our Work:} As a representative of infinitely regular wavelets, ``dmey" serves as a point of comparison against wavelets with finite support.
\end{itemize}

This overview illustrates the rich diversity within wavelet families. Our empirical results suggest a strong correlation between these properties (\eg, smoothness, support length) and their suitability for transferring knowledge across different neural network architectures and scaling directions.

\begin{table}[ht]
\centering
\caption{The structures of DeiT models used in our experiments.}
\label{tab:deit_structures}
\resizebox{0.4\textwidth}{!}{
\begin{tabular}{@{}lcccc@{}}
\toprule
\textbf{Config} & \textbf{DeiT-Ti} & \textbf{DeiT-S} & \textbf{DeiT-B} & \textbf{DeiT-L} \\
\midrule
\# layers      & 3          & 6           & 12          & 24                \\
\# hidden       & 192         & 384         & 768         & 1024               \\
\# heads        & 3           & 6           & 12           & 16                 \\
input size    & 224         & 224         & 224         & 224               \\
patch size    & 16          & 16         & 16          & 16                  \\
\bottomrule
\end{tabular}
}
\end{table}

\begin{table}
    \centering
    \caption{Hyper-parameters for all baseline on ImageNet-1K.}
    \vspace{-0.1in}
    \setlength{\tabcolsep}{4.6 mm}
        \begin{tabular}{@{}lr@{}}
        \toprule[1.3pt]
        \textbf{Training Settings} & \textbf{Configuration} \\
        \midrule[1.1pt]
        optimizer & AdamW\\
        base learning rate &  S: 2.5e-4 $\mid$ B: 1.25e-4\\
        warmup learning rate & 1e-6\\
        weight decay & 0.05\\
        optimizer momentum & 0.9\\
        batch size & S: 256 $\mid$ B: 128\\
        training epochs & 150\\
        learning rate schedule & cosine decay\\
        warmup epochs & 0\\
        color jitter & 0.4 \\
        auto augment & rand-m9-mstd0.5-inc1\\
        mixup & 0.8\\
        cutmix & 1.0\\
        label smoothing & 0.1\\
        drop path & 0.1\\
        \bottomrule[1.3pt]
        \end{tabular}
    \label{tab:hyper_image1k}
    \vspace{-0.05in}
\end{table}

\begin{table}
    \centering
    \setlength{\tabcolsep}{1.3 mm}
    \caption{Characteristics of downstream datasets.}
    \vspace{-0.1in}
    \resizebox{0.48\textwidth}{!}{
        \begin{tabular}{@{}lcccc@{}}
        \toprule[1.3pt]
        \textbf{Dataset} & \textbf{Classes} & \textbf{Total} & \textbf{Training} & \textbf{Testing} \\
        \cmidrule[1.1pt]{1-5}
        \textbf{Oxford Flowers}~\cite{nilsback2008automated} & 102 & 8,189 & 2,040  & 6,149\\
        \textbf{CUB-200-2011}~\cite{wah2011caltech} & 200 & 11,788 & 5,994 & 5,794 \\
        \textbf{Stanford Cars}~\cite{gebru2017fine} & 196 & 16,185 & 8,144 & 8,041\\
        \textbf{CIFAR10}~\cite{krizhevsky09} & 10 & 60,000 & 50,000 & 10,000 \\
        \textbf{CIFAR100}~\cite{krizhevsky09} & 100 & 60,000 & 50,000 & 10,000 \\
        \textbf{Food101}~\cite{bossard2014food} & 101 & 101,000 & 75,750 & 25,250\\
        \textbf{iNat-2019}~\cite{tan2019herbarium} & 1010 & 268,243 & 265,213 & 3,030\\
        \bottomrule[1.3pt]
        \end{tabular}
        }
    \label{tab:datasets}
\end{table}

\begin{table*}
    \centering
    \setlength{\tabcolsep}{0.7 mm}
    \caption{Hyper-parameters for neural networks trained on downstream datasets.}
    \vspace{-0.1in}
    \resizebox{\textwidth}{!}{
        \begin{tabular}{@{}lccccccccccccc@{}}
        \toprule[1.3pt]
        \textbf{Dataset} & \makecell{\textbf{Batch}\\ \textbf{Size}} & \makecell{\textbf{Epoch}} & \makecell{\textbf{Learning}\\ \textbf{Rate}} & \makecell{\textbf{Drop}\\ \textbf{Last}} & \makecell{\textbf{Warmup}\\ \textbf{Epochs}} & \makecell{\textbf{Droppath}\\ \textbf{Rate}} & \makecell{\textbf{Color}\\ \textbf{Jitter}} & \makecell{\textbf{Auto} \\ \textbf{Augment}} & \makecell{\textbf{Random}\\ \textbf{Rrase}} & \makecell{\textbf{Mixup}} & \makecell{\textbf{Cutmix}} & \makecell{\textbf{Scheduler}} & \makecell{\textbf{Optimizer}}\\
        \midrule[1.1pt]
        \textbf{Oxford Flowers} & 512 & 300 & 3e-4 & False & 0 & 0 & 0.4 & \multirow{7}{*}{\rotatebox{90}{\fontsize{8.5}{12}\selectfont rand-m9-mstd0.5-inc1}} & 0.25 & 0 & 0 & cosine & AdamW \\
        \textbf{CUB-200-2011} & 512 & 300 & 3e-4 & False & 0 & 0.1 & 0 & & 0.25 & 0 & 0 & cosine & AdamW \\
        \textbf{Stanford Cars} & 512 & 300 & 3e-4 & False & 0 & 0.1 & 0 & & 0.25 & 0 & 0 & cosine & AdamW \\
        \textbf{CIFAR10} & 512 & 300 & 5e-4 & True & 0 & 0.1 & 0.4 & & 0.25 & 0 & 0 & cosine & AdamW\\
        \textbf{CIFAR100} & 512 & 300 & 5e-4 & True & 0 & 0.1 & 0.4 & & 0.25 & 0 & 0 & cosine & AdamW\\
        \textbf{Food101} & 512 & 300 & 5e-4 & True & 0 & 0.1 & 0.4 & & 0.25 & 0 & 0 & cosine & AdamW\\
        \textbf{iNat-2019} & 512 & 100 & 5e-4 & True & 0 & 0.1 & 0.4 & & 0.25 & 0 & 0 & cosine & AdamW\\
        \bottomrule[1.3pt]
        \end{tabular}
        }
    \label{tab:hyper_down}
\end{table*}

\begin{table}[ht]
\centering
\caption{The structures of BERT models. }
\label{tab:bert_structures}
\resizebox{0.4\textwidth}{!}{
\begin{tabular}{@{}lcccc@{}}
\toprule
\textbf{Config} &\textbf{BERT-Ti} &  \textbf{BERT-S} & \textbf{BERT-B} & \textbf{BERT-L}  \\
\midrule
\# layers  &3    & 6          & 12          & 24                  \\
\# hidden &192      & 384         & 768         & 1024                 \\
\# heads  &3      & 6           & 12          & 16                      \\
\# vocab  & 30522       & 30522       & 30522       & 30522               \\
seq. length & 128  & 128         & 128         & 128                  \\
\bottomrule
\end{tabular}
}
\end{table}

\begin{table}[ht]
\centering
\caption{Hyper-parameters for BERT and RoBERTa pre-training.}
\label{tab:hyper_bert}
\setlength{\tabcolsep}{5mm} 
\resizebox{0.35\textwidth}{!}{
\begin{tabular}{@{}ll@{}}
\toprule[1.3pt]
\textbf{Training Setting} & \textbf{Configuration} \\
\midrule[1.1pt]
Optimizer & AdamW \\
Adam $\epsilon$ & 1e-6 \\
Learning Rate & 2e-4 \\
Weight Decay & 0.1 \\
Learning Rate Schedule & Linear Decay \\
Warmup Steps & 0 \\
Total Training Steps & 400,000 \\
Batch Size & 256 \\
Sequence Length & 128 \\
Dataset & English Wikipedia \\
\bottomrule[1.3pt]
\end{tabular}}
\end{table}

\begin{table}[ht]
\centering
\caption{The structures of RoBERTa and GPT-2 models used in our experiments. RoBERTa shares a similar architecture to BERT but uses a different vocabulary.}
\label{tab:gpt2_structures}
\resizebox{0.4\textwidth}{!}{
\begin{tabular}{@{}lcccc@{}}
\toprule
\textbf{Config} &\textbf{RoBERTa-S} & \textbf{RoBERTa-B}& \textbf{GPT-S} & \textbf{GPT-B}   \\
\midrule
\# layers      & 6              & 12    & 6              & 12              \\
\# hidden       & 384             & 768   & 384            & 768            \\
\# heads        & 6               & 12     & 6             & 12         \\
\# vocab    & 50265          & 50265      & 50257           & 50257          \\
seq. length    & 128            & 128  & 1024            & 1024           \\
\bottomrule
\end{tabular}}
\end{table}

\begin{table}[ht]
\centering
\caption{Hyper-parameters for GPT-2 pre-training.}
\label{tab:hyper_gpt}
\setlength{\tabcolsep}{5mm} 
\resizebox{0.4\textwidth}{!}{
\begin{tabular}{@{}ll@{}}
\toprule[1.3pt]
\textbf{Training Setting} & \textbf{Configuration} \\
\midrule[1.1pt]
Optimizer & AdamW \\
Adam $\epsilon$ & 1e-8 \\
Adam $\beta_1, \beta_2$ & 0.9, 0.95 \\
Learning Rate &  1e-3 \\
Weight Decay & 0.1 \\
Learning Rate Schedule & Cosine Decay \\
Warmup Steps & 0 \\
Total Training Steps &  S: 150000 $\mid$ B: 50000 \\
Batch Size & 512 \\
Sequence Length & 1024 \\
Dropout & 0.1 \\
Dataset & Wikipedia + Book Corpus \\
\bottomrule[1.3pt]
\end{tabular}
}
\end{table}


    

\section{Training Details}
\label{app:sec_training_details}

\subsection{Vision Models Training}
\label{app:subsec_vision_details}

\subsubsection{Experiment Structures.}

We show DeiT structures in Table~\ref{tab:deit_structures}.

\subsubsection{Details of Scaling Experiments on DeiT}
\label{subsubsec:scaling_details}

To evaluate the effectiveness of our bidirectional transfer method, BoT, we conducted scaling experiments on the ImageNet-1K dataset using DeiT architectures. The experiments cover both L2S and S2L scenarios.

For the L2S transfer, we initialized a DeiT-S model using the weights from a pre-trained DeiT-B. For the S2L transfer, we initialized a DeiT-B model using a pre-trained DeiT-S. These initialized models, along with all baselines, were then pre-trained for 150 epochs. The key hyper-parameters for this pre-training phase are detailed in Table~\ref{tab:hyper_image1k}. Note that distinct learning rates and batch sizes were used for the DeiT-S and DeiT-B to ensure stable and optimal training for each architecture. All other settings, including data augmentation strategies like Mixup and Cutmix, remained consistent across experiments to allow for a fair comparison of the initialization methods.

\subsubsection{Details of Downstream Datasets}
\label{app:dataset}
Additional datasets include Oxford Flowers~\cite{nilsback2008automated}, CUB-200-2011~\cite{wah2011caltech}, Stanford Cars~\cite{gebru2017fine}, CIFAR-10, CIFAR-100~\cite{krizhevsky09}, Food-101~\cite{bossard2014food}, and iNaturalist-2019~\cite{tan2019herbarium}.
Table~\ref{tab:datasets} presents the details of seven downstream datasets, which are sorted by the size of datasets. Table~\ref{tab:hyper_down} presents the basic settings, including batch size, warmup epochs, training epochs and other settings for training the models initialized with all baselines on various datasets.

\subsection{Bert Training}
\label{app:subsec_bert}

\subsubsection{Experiment Structures.}
We show BERT structures in Table~\ref{tab:bert_structures}.

\subsubsection{Details of Scaling Experiments on BERT}
\label{subsubsec:scaling_details_bert}

To validate BoT's training acceleration on BERT~\cite{devlin2018bert}, we conducted bidirectional pre-training experiments on the English Wikipedia corpus. The experiments covered both L2S and S2L transfers. For the L2S scenario, we initialized a BERT-S model using a pre-trained BERT-B. Conversely, for the S2L scenario, a BERT-B model was initialized from a pre-trained BERT-S. All initialized models, along with their respective baselines, were then pre-trained for 400K steps. The key hyper-parameters for this pre-training phase are detailed in Table~\ref{tab:hyper_bert}. A consistent batch size of 256 was used for all runs.

\subsubsection{Details of Downstream Task Fine-tuning}
\label{subsubsec:downstream_nlp_details}

To assess the quality of the transferred knowledge, all initialized models were subsequently evaluated on the GLUE~\cite{wang2018glue} and SQuAD~\cite{rajpurkar2016squad, rajpurkar2018know} benchmarks without any intermediate pre-training.

For the GLUE benchmark, we performed a hyper-parameter search for each task. We set the batch size to 32 and the learning rate to $1\text{e-}4$ and the optimal number of training epochs from $\{3, 6, 10\}$. We used the Adam optimizer~\cite{kingma2014adam} for all runs and report accuracy for most tasks, with the exception of CoLA, for which we report the Matthews Correlation Coefficient (MCC), and STS-B, for which we report Pearson correlation.

For the SQuAD (v1.1 and v2.0) fine-tuning, we used a fixed setting with a batch size of 12, a learning rate of 3e-5, and training epochs from $\{2, 4, 8\}$. Performance is measured by the standard Exact Match (EM) and F1 scores.

For all tasks, we report the final metric on the development set, averaged over 5 independent runs with different random seeds to ensure robust and reproducible results. An average score is also computed for each benchmark suite.

\subsection{RoBerta Training}
\label{app:subsec_roberta}
\subsubsection{Experiment Structures.}
We show RoBERTa structures in Table~\ref{tab:gpt2_structures}.

\subsubsection{Details of Scaling Experiments on RoBERTa}
\label{subsubsec:scaling_details_roberta}

To validate BoT's training acceleration on RoBERTa~\cite{liu2019roberta}, we conducted bidirectional pre-training experiments on the English Wikipedia corpus. The experiments covered both L2S and S2L transfers. The key hyper-parameters for this pre-training phase are detailed in Table~\ref{tab:hyper_bert}.

\subsection{GPT2 Training}
\label{app:subsec_gpt2}
\subsubsection{Experiment Structures.}
We show GPT-2 structures in Table~\ref{tab:gpt2_structures}.

\subsubsection{Details of Scaling Experiments on GPT}

To evaluate the performance of our method, BoT, on autoregressive models, we performed both L2S and S2L transfers. For the L2S scenario, we initialized a GPT-Small model using a pre-trained GPT-Base. Conversely, for the S2L scenario, a GPT-Base model was initialized from a pre-trained GPT-Small. The initialized models, along with baselines, were then pre-trained on a dataset combining the English Wikipedia corpus and the Toronto Book Corpus~\cite{zhu2015aligning}. The total training duration was 200K steps. The specific hyper-parameters used for this pre-training phase are detailed in Table~\ref{tab:hyper_gpt}.

\subsection{LiGO and Mango}
In experiments of DeiTs, we train LiGO and Mango operators with AdamW optimizer for 100 steps on ImageNet-1K. The learning rate is 5e-4. Batch size is 256. In experiments of BERT and RoBERTa models, we train LiGO and Mango operators with Adamw optimizer for 100 steps on the concatenation of English Wikipedia. The learning rate is 2e-4. Minibatch is 256. In experiments of the GPT models, we train LiGO and Mango operators with Adamw optimizer for 100 steps on the concatenation of English Wikipedia and Toronto Book Corpus. The learning rate is 1e-5. Minibatch is 512.

\section{Additional Results}
\label{app:sec_additional_results}

    


\begin{table*}[t]
\centering
\caption{Performance comparison on the GLUE and SQuAD benchmarks for RoBERTa. Models are initialized and then directly fine-tuned without additional pre-training. The ``Avg'' columns show the average scores for each benchmark. }
\label{tab:downstream_roberta}
\setlength{\tabcolsep}{3pt} 
\resizebox{0.95\textwidth}{!}{
\begin{tabular}{@{}cl ccccccc>{\columncolor{gray!15}}c cccc>{\columncolor{gray!15}}c@{}}
\toprule[1.5pt]
\multirow{2}{*}{Model} & \multirow{2}{*}{Methods} & \multicolumn{8}{c}{GLUE Benchmark} & \multicolumn{5}{c}{SQuAD Benchmark} \\
\cmidrule(lr){3-10} \cmidrule(lr){11-15}
& & SST-2 & MNLI & MRPC & CoLA & QNLI & QQP & STS-B & \textbf{Avg} & v1.1 (EM) & v1.1 (F1) & v2.0 (EM) & v2.0 (F1) & \textbf{Avg} \\
\midrule

\multirow{3}{*}{\rotatebox{90}{\scriptsize \textbf{RoBERTa-S}}}
& Scratch & 88.19 & 75.30 & 76.47 & 21.27 & 85.19 & 88.48 & 82.87 & 73.97 & 7.99 & 18.02 & 3.39 & 8.16 & 9.39 \\
& WS & 88.42 & 75.26 & 78.43 & 25.04 & 83.53 & 88.12 & 80.84 & 74.23 & 23.75 & 32.72 & 12.79 & 18.04 & 21.83 \\
& \cellcolor{blue!12}\textbf{DWT} & \cellcolor{blue!12}\textbf{88.42} & \cellcolor{blue!12}\textbf{76.08} & \cellcolor{blue!12}\textbf{78.43} & \cellcolor{blue!12}\textbf{23.27} & \cellcolor{blue!12}\textbf{85.58} & \cellcolor{blue!12}\textbf{88.62} & \cellcolor{blue!12}\textbf{82.81} & \cellcolor{blue!12}\textbf{74.74} & \cellcolor{blue!12}\textbf{42.58} & \cellcolor{blue!12}\textbf{54.22} & \cellcolor{blue!12}\textbf{20.26} & \cellcolor{blue!12}\textbf{26.87} & \cellcolor{blue!12}\textbf{35.98} \\
\midrule

\multirow{5}{*}{\rotatebox{90}{\scriptsize{\textbf{RoBERTa-B}}}}
& Scratch & 81.65 & 31.82 & 69.61 & 2.16 & 61.58 & 77.50 & 15.69 & 48.57 & 8.80 & 18.76 & 3.92 & 8.86 & 10.09 \\
& bert2BERT & 81.19 & 61.91 & 70.59 & 10.51 & 59.86 & 79.94 & 20.86 & 54.98 & 13.64 & 24.08 & 5.39 & 10.62 & 13.43 \\
& LiGO & 87.61 & 76.87 & 71.08 & 21.72 & 83.40 & 87.83 & 70.86 & 71.34 & 68.34 & 78.41 & 33.58 & 38.84 & 54.79 \\
& Mango & 87.50 & 77.04 & 70.59 & 21.72 & 83.34 & 87.79 & 69.46 & 71.06 & 68.75 & 78.63 & 33.53 & 38.90 & 54.95 \\
& \cellcolor{blue!12}\textbf{DWT} & \cellcolor{blue!12}\textbf{88.92} & \cellcolor{blue!12}\textbf{78.82} & \cellcolor{blue!12}\textbf{72.38} & \cellcolor{blue!12}\textbf{22.45} & \cellcolor{blue!12}\textbf{84.54} & \cellcolor{blue!12}\textbf{87.91} & \cellcolor{blue!12}\textbf{71.12} & \cellcolor{blue!12}\textbf{72.31} & \cellcolor{blue!12}\textbf{69.87} & \cellcolor{blue!12}\textbf{79.18} & \cellcolor{blue!12}\textbf{34.22} & \cellcolor{blue!12}\textbf{39.72} & \cellcolor{blue!12}\textbf{55.75} \\

\bottomrule[1.5pt]
\end{tabular}%
}
\end{table*}

\begin{figure*}[ht]
\vspace{-0.14in}
    \centering
    \includegraphics[width=\linewidth]{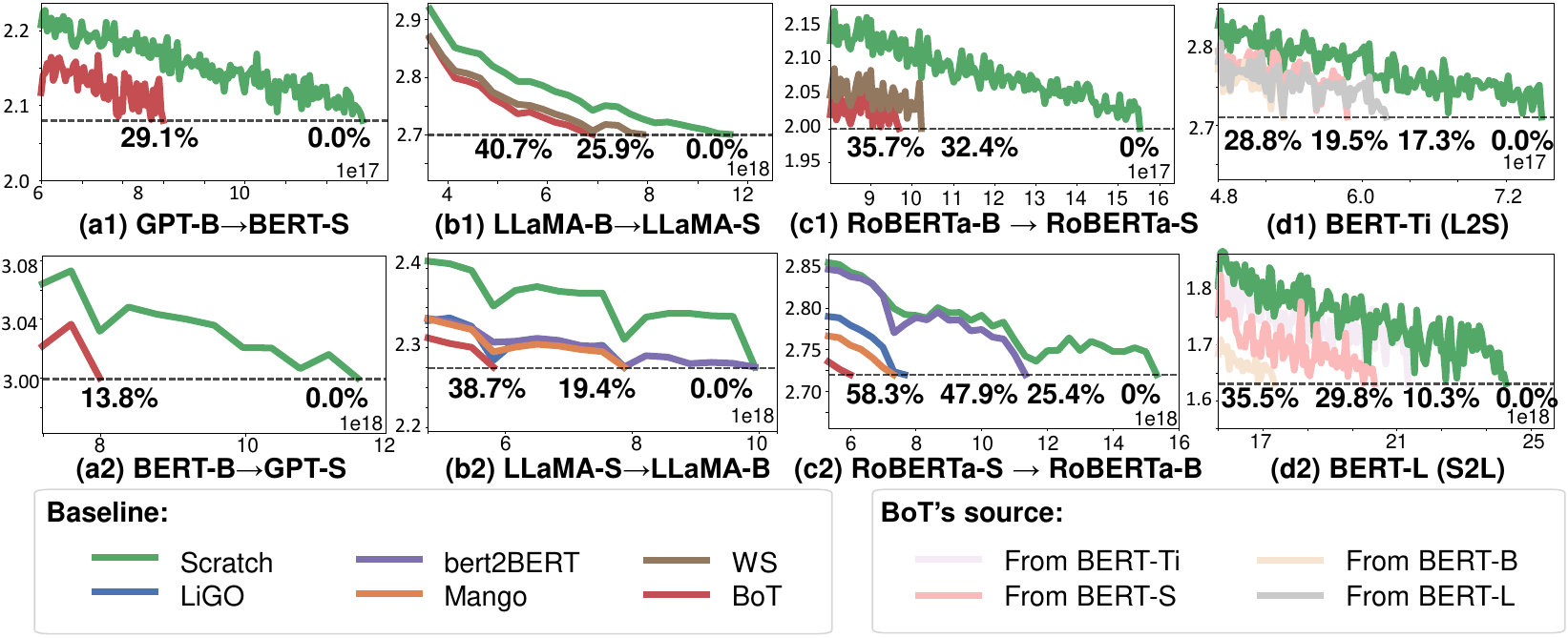}
    \vspace{-0.27in}
    \caption{(a) BERT$\leftrightarrow$GPT; (b) Generalization to LLaMA; (c) Generalization to RoBERTa; (d) Scalability on Tiny and Large.}
    \label{fig:app_results}
    
\end{figure*}



\subsection{Cross-Architecture Knowledge Transfer}
\label{subsec:cross_arch_transfer}

To evaluate the versatility of the BoT framework across fundamentally different model architectures, we conducted bidirectional transfer experiments between encoder-dominated and decoder-dominated Transformers. Specifically, we initialized a BERT-S model from a GPT-B model and, conversely, a GPT-S model from a BERT-B model. As shown in Figure~\ref{fig:app_results}(a), BoT successfully bridges the architectural gap between these differing paradigms. In the GPT-B $\rightarrow$ BERT-S setting, BoT achieves a significant 29.1\% reduction in FLOPs, demonstrating that even under conditions of architectural mismatch, our framework can still identify and effectively transfer critical knowledge. Similarly, in the BERT-B $\rightarrow$ GPT-S direction, BoT accelerates convergence with a 13.8\% FLOPs saving. These results highlight the robust generalization capabilities of our approach, proving its efficacy well beyond homologous architectural families.

\subsection{Results on LLaMA}
\label{subsec:llama_results}

To further assess the generality of our framework on modern large language models, we conducted parallel experiments on the LLaMA architecture. Following the established protocol, we performed bidirectional pre-training by initializing a LLaMA-S from a LLaMA-B and vice versa. The pre-training convergence curves, illustrated in Figure~\ref{fig:app_results}(b), affirm BoT's robust performance on decoder-only generative models. In the L2S scenario (LLaMA-B $\rightarrow$ LLaMA-S), BoT achieves a 40.7\% FLOPs saving, ensuring rapid adaptation to a smaller model. This strong performance extends to the S2L direction (LLaMA-S $\rightarrow$ LLaMA-B), where our approach delivers a 38.7\% FLOPs saving, significantly outperforming alternative expansion baselines. These findings indicate that our framework effectively scales to state-of-the-art LLM architectures, facilitating highly efficient bidirectional knowledge transfer.

\subsection{Results on RoBERTa}
\label{app:subsec_roberta_results}

To further assess the generality of our framework, we conducted parallel experiments on RoBERTa~\cite{liu2019roberta}. Following the same protocol, we performed bidirectional pre-training on the English Wikipedia corpus, initializing a RoBERTa-S from a RoBERTa-B and vice versa. The initialized models were subsequently evaluated on the GLUE and SQuAD benchmarks to measure the quality of transferred knowledge.

The pre-training convergence curves for RoBERTa, illustrated in Figure~\ref{fig:app_results}(c), affirm BoT's robust, state-of-the-art performance. In the L2S scenario, BoT achieves the fastest convergence, yielding a 35.7\% FLOPs saving that significantly outperforms both KD and WS. This strong performance is mirrored in the S2L direction, where our approach again leads all baselines, delivering a 54.7\% FLOPs saving, a massive +32.6\% improvement over the expansion methods, Mango and LiGO. These results demonstrate that our framework's efficiency gains are not specific to one architecture but are broadly applicable across different Transformer variants.

The downstream task results, presented in Table~\ref{tab:downstream_roberta}, further validate the superiority of our initialization. In the L2S setting for RoBERTa-S, BoT achieves the highest average scores on both GLUE (74.74) and SQuAD (35.98), dramatically outperforming the WS baseline, especially on SQuAD (+14.15 points). In the S2L setting for RoBERTa-B, BoT again secures the top performance on both benchmarks, surpassing all contemporary expansion methods including LiGO and Mango. This consistent, state-of-the-art performance across different model and scaling directions solidifies our framework as a powerful and universally effective solution for cross-architecture knowledge transfer.

\begin{table*}[t]
\centering
\caption{Performance comparison on GLUE and SQuAD benchmarks for RoBERTa after pre-training from initialization. This table shows the final performance of models that were first pre-trained and then fine-tuned. ``Avg'' columns show the average scores for each benchmark.}
\label{tab:post_pretraining_roberta}
\setlength{\tabcolsep}{2.5pt} 
\resizebox{0.95\textwidth}{!}{
\begin{tabular}{@{}cl cc ccccccc>{\columncolor{gray!15}}c cc>{\columncolor{gray!15}}c@{}}
\toprule[1.5pt]
\multirow{2}{*}{Model} & \multirow{2}{*}{Methods} & \multicolumn{2}{c}{Savings} & \multicolumn{8}{c}{GLUE Benchmark} & \multicolumn{3}{c}{SQuAD Benchmark} \\
\cmidrule(lr){3-4} \cmidrule(lr){5-12} \cmidrule(lr){13-15}
& & FLOPs & Walltime & SST-2 & MNLI & MRPC & CoLA & QNLI & QQP & STS-B & \textbf{Avg} & v1.1 (f1/em) & v2.0 (f1/em) & \textbf{Avg} \\
\midrule

\multirow{3}{*}{\rotatebox{90}{\scriptsize{\textbf{RoBERTa-S}}}}
& Scratch & 0.0\% & 0.0\% & 89.56 & 76.64 & 79.41 & 20.39 & 84.90 & 88.32 & 82.59 & 74.54 & 76.2 / 65.8 & 37.5 / 32.0 & 52.9 \\
& WS & 32.4\% & 32.3\% & 88.50 & 76.85 & 78.19 & 21.62 & 83.54 & 88.15 & 81.43 & 74.04 & 75.3 / 64.7 & 37.6 / 32.1 & 52.4 \\

& \cellcolor{blue!12}\textbf{BoT} & \cellcolor{blue!12}\textbf{35.7\%} & \cellcolor{blue!12}\textbf{35.8\%} & \cellcolor{blue!12}\textbf{89.33} & \cellcolor{blue!12}\textbf{76.78} & \cellcolor{blue!12}\textbf{76.47} & \cellcolor{blue!12}\textbf{23.18} & \cellcolor{blue!12}\textbf{84.61} & \cellcolor{blue!12}\textbf{88.71} & \cellcolor{blue!12}\textbf{82.93} & \cellcolor{blue!12}\textbf{74.57} & \cellcolor{blue!12}\textbf{76.7 / 66.3} & \cellcolor{blue!12}\textbf{37.9 / 32.4} & \cellcolor{blue!12}\textbf{53.3} \\
\midrule

\multirow{5}{*}{\rotatebox{90}{\scriptsize{\textbf{RoBERTa-B}}}}
& Scratch & 0.0\% & 0.0\% & 90.71 & 82.59 & 83.33 & 45.87 & 89.75 & 90.38 & 85.44 & 81.15 & 84.4 / 74.7 & 43.6 / 38.8 & 60.4 \\
& bert2BERT & 7.2\% & 6.6\% & 91.63 & 82.00 & 86.03 & 41.04 & 89.00 & 90.21 & 85.20 & 80.73 & 83.8 / 74.2 & 41.5 / 36.6 & 59.0 \\
& LiGO & 22.1\% & 20.3\% & 90.37 & 82.38 & 83.82 & 44.96 & 90.19 & 90.18 & 85.16 & 81.01 & 84.7 / 75.5 & 42.3 / 37.3 & 59.9 \\
& Mango & 22.2\% & 20.4\% & 90.83 & 81.99 & 84.31 & 44.36 & 90.43 & 90.23 & 85.87 & 81.15 & 84.6 / 76.3 & 42.2 / 38.4 & 60.4 \\

& \cellcolor{blue!12}\textbf{BoT} & \cellcolor{blue!12}\textbf{54.7\%} & \cellcolor{blue!12}\textbf{51.5\%} & \cellcolor{blue!12}\textbf{90.60} & \cellcolor{blue!12}\textbf{82.91} & \cellcolor{blue!12}\textbf{84.82} & \cellcolor{blue!12}\textbf{44.00} & \cellcolor{blue!12}\textbf{90.13} & \cellcolor{blue!12}\textbf{90.19} & \cellcolor{blue!12}\textbf{85.70} & \cellcolor{blue!12}\textbf{81.19} & \cellcolor{blue!12}\textbf{84.7 / 74.2} & \cellcolor{blue!12}\textbf{43.8 / 39.2} & \cellcolor{blue!12}\textbf{60.5} \\
\bottomrule[1.5pt]
\end{tabular}%
}
\end{table*}

To further assess the generality of our framework, we conducted parallel experiments on RoBERTa~\cite{liu2019roberta}. We evaluated bidirectional transfers (RoBERTa-B $\to$ RoBERTa-S and vice versa) by first pre-training the initialized models and then fine-tuning them on the GLUE and SQuAD benchmarks. The results are presented in Table~\ref{tab:post_pretraining_roberta}. In both L2S and S2L scenarios, BoT once again demonstrates unparalleled computational efficiency. On RoBERTa-S, BoT leads in efficiency with a 35.7\% FLOPs saving, surpassing the standard WS baseline. Similarly, on RoBERTa-B, BoT achieves a state-of-the-art 54.7\% FLOPs saving and 51.5\% Walltime saving, which is over 30 percentage points higher than the methods, LiGO and Mango. While delivering this massive efficiency gain, its final downstream performance on GLUE (81.19) and SQuAD (60.5) remains highly competitive and on par with these more resource-intensive baselines. In contrast, BoT provides strong downstream performance while strictly reducing training costs. These results on RoBERTa reinforce our primary finding: BoT achieves both high computational efficiency and strong final model performance, making it a highly practical solution for cross-architecture knowledge transfer.

\subsection{Impact of Extreme Scale Disparity}
\label{subsec:extreme_scale_disparity}

To analyze how scale disparities between source and target models affect BoT's performance, we conducted two sets of bidirectional experiments. First, in an L2S setting, we initialized a BERT-Ti model from a BERT-S, a BERT-B, and a much larger BERT-L. Second, in an S2L setting, we initialized a BERT-L model from a BERT-B, a BERT-S, and an extremely small BERT-Ti. We compared the convergence of all initialized models against a standard from-scratch baseline. The results, illustrated in Figure~\ref{fig:app_results}(d), reveal key insights into the trade-off between knowledge capacity and scale disparity under severe size mismatches.

In the L2S scenario (transferring to BERT-Ti), the results reveal a delicate trade-off between the source model's knowledge capacity and the architectural scale gap. As seen in Figure~\ref{fig:app_results}(d1), initializing from BERT-B achieves the fastest convergence with a 28.8\% FLOPs saving, outperforming BERT-S which yields a 19.5\% saving. This indicates that despite a larger size gap, the richer knowledge within the BERT-B architecture provides superior transferable benefits. However, when the source model is scaled up further to the massive BERT-L, the FLOPs saving drops to 17.3\%. This suggests that an overly extreme scale disparity can hinder the effective mapping of features, eventually outweighing the advantages of a larger knowledge base. 

Conversely, in the S2L scenario, a larger source model consistently provides a fundamentally more potent initialization. As shown in Figure~\ref{fig:app_results}(d2), initializing from the adjacent BERT-B is highly effective, achieving a 35.5\% FLOPs saving. This is followed by BERT-S with a 29.8\% saving, while initializing from the extremely small BERT-Ti yields only a 10.3\% FLOPs saving. In this expansion direction, a larger source model inherently possesses a richer knowledge base, which consequently translates to a smaller scale disparity relative to the target model. Therefore, having more knowledge directly correlates with a reduced structural gap, leading to markedly better transfer performance.

Collectively, these extreme bidirectional experiments demonstrate how BoT navigates the complexities of size mismatch. For adaptation to a tiny target (L2S), optimal efficiency requires balancing knowledge richness with structural similarity, where a moderately larger source (e.g., BERT-B) serves as the ideal teacher. For extreme model expansion (S2L), maximizing the source model's size simultaneously minimizes the scale gap and provides the most comprehensive foundational knowledge, making it unequivocally advantageous.

\begin{table*}[ht]
\centering
\setlength{\tabcolsep}{6pt}
\begin{tabular}{lcccccccc}
\toprule
Method              & CIFAR-10 & CIFAR-100 & CUB200 & Flowers102 & Food-101 & iNat-2019 & Cars & Avg. \\
\midrule
Random Uniform             & 95.13    & 74.73     & 62.53  & 69.01      & 73.94    & 51.92     & 52.11 & 68.85 \\
Random Gaussian     & 96.87    & 80.14     & 73.91  & 90.63      & 83.19    & 65.02     & 79.92 & 81.67 \\
Zero Padding (BoT)  & \textbf{97.24} & \textbf{82.38} & \textbf{74.08} & \textbf{94.71} & \textbf{84.44} & \textbf{67.14} & \textbf{88.64} & \textbf{84.37} \\
\bottomrule
\end{tabular}
\vspace{-0.5em}
\caption{Ablation on padding strategies for high-frequency coefficients in IDWT-based S2L transfer. Zero padding consistently yields the best average accuracy. We report top-1 accuracy (\%) on seven downstream datasets after direct fine-tuning.}
\label{tab:padding-ablation}
\end{table*}

\subsection{Ablation on High-Frequency Coefficient Padding Strategies in S2L Transfer}
\label{sec:appendix-padding}

To further investigate the role of high-frequency detail coefficients in \textbf{Small-to-Large (S2L)} transfers, we conduct an ablation study comparing different padding strategies when reconstructing the target base model from a smaller source model using IDWT. In our framework (BoT), the high-frequency sub-bands $\{cD_m\}_{m=1}^7$ are typically set to zero tensors (zero padding). We hypothesize that the initialization distribution of these coefficients may influence convergence and downstream transferability.

We perform S2L transfer from a DeiT-S (6-layer, hidden size 384) to a DeiT-B (12-layer, hidden size 768). We compare three strategies for initializing the high-frequency detail coefficients before applying 3D-IDWT:
\begin{enumerate}
\item \textbf{Random Uniform Padding}: coefficients are sampled from $\mathcal{U}(-\sigma, \sigma)$ with $\sigma$ defined as above. (Similar to Gaussian padding; omitted in main table.)

  \item \textbf{Random Gaussian Padding}: all $cD_m$ sub-bands are filled with values drawn from $\mathcal{N}(0, \sigma^2)$, where $\sigma$ matches the empirical standard deviation of the corresponding low-frequency coefficients $cA$ in the source model.
  
    \item \textbf{Zero Padding (default BoT)}: all $cD_m$ sub-bands are set to zero tensors.

\end{enumerate}

From Table~\ref{tab:padding-ablation}, zero padding consistently outperforms other padding across all datasets---yielding a $+2.70\%$ average accuracy improvement.
This confirms that setting high-frequency coefficients to zero provides a cleaner and more stable initialization for IDWT-based reconstruction, avoiding the noise introduced by randomly sampled detail bands.
Although Gaussian padding still significantly outperforms training from scratch, our results suggest that such random detail initialization may misalign with the structural correlations embedded in the learngene, thereby hindering fine-grained recognition tasks (e.g., Cars, CUB200). In practice, \textbf{zero padding remains the most effective and robust strategy}, reinforcing our choice in the main BoT pipeline.

\end{document}